\PassOptionsToPackage{table,dvipsnames}{xcolor}

\documentclass[11pt]{article} 
\usepackage[preprint]{acl} 

\usepackage{times} 
\usepackage{latexsym}

\usepackage[T1]{fontenc}

\usepackage[utf8]{inputenc}

\usepackage{microtype}

\usepackage{inconsolata}

\usepackage{graphicx}

\usepackage{amsmath}
\usepackage{amsfonts}

\usepackage{cleveref}
\usepackage{stmaryrd}
\usepackage{bbm}

\usepackage{booktabs}
\usepackage{multirow}
\usepackage[table]{xcolor}
\usepackage{siunitx}

\definecolor{blue-violet}{rgb}{0.54, 0.17, 0.89}
\definecolor{nicegreen}{RGB}{102, 204, 102}
\definecolor{niceorange}{RGB}{255,140,0}
\definecolor{niceyellow}{RGB}{255,215,0}
\definecolor{good}{RGB}{220,240,220}
\definecolor{warn}{RGB}{255,245,210}
\definecolor{bad}{RGB}{255,225,200}

\usepackage{fixmetodonotes}

\usepackage{enumitem} 
\usepackage{titletoc} 

\usepackage{xspace}
\newcommand{\ours}{DRIV-EX\xspace}

\usepackage[ruled,vlined]{algorithm2e} 

\title{\ours: Counterfactual Explanations for Driving LLMs}

\author{
 \textbf{Amaia Cardiel\textsuperscript{1,2}},
 \textbf{Éloi Zablocki\textsuperscript{2}},
 \textbf{Elias Ramzi\textsuperscript{2}},
 \textbf{Eric Gaussier\textsuperscript{1}}
\\
\\
 \textsuperscript{1} Aptikal, Université Grenoble Alpes, Grenoble, France \\
 \textsuperscript{2} Valeo.ai, Paris, France
}

\begin{document}

\maketitle

\begin{abstract}

Large language models (LLMs) are increasingly used as reasoning engines in autonomous driving, yet their decision-making remains opaque. We propose to study their decision process through counterfactual explanations, which identify the minimal semantic changes to a scene description required to alter a driving plan.
We introduce \ours, a method that leverages gradient-based optimization on continuous embeddings to identify the input shifts required to flip the model's decision. Crucially, to avoid the incoherent text typical of unconstrained continuous optimization, \ours uses these optimized embeddings solely as a \emph{semantic guide}: they are used to bias a controlled decoding process that re-generates the original scene description. This approach effectively steers the generation toward the counterfactual target while guaranteeing the linguistic fluency, domain validity, and proximity to the original input essential for interpretability.
Evaluated using the LC-LLM planner on the textual highD dataset, \ours generates valid, fluent counterfactuals more reliably than existing baselines. It successfully exposes latent biases and provides concrete insights to improve the robustness of LLM-based driving agents. The code is available at \url{https://github.com/Amaia-CARDIEL/DRIV_EX}.

\end{abstract}

\section{Introduction}
\label{sec:introduction}

End-to-end Autonomous Driving (AD) increasingly relies on Large Language Models (LLMs) and Vision-Language Models (VLMs). Thanks to large-scale pre-training, these models can integrate user instructions and sensor inputs to directly generate driving decisions \citep{rowe2025poutine,hwang2025emma,renz2024carllava,renz2025simlingo,nvidia2025alpamayo}.
Unlike traditional neural planners that operate as black boxes, LLM-based systems provide natural language explanations, e.g., with Chain-of-Thought (CoT), often structured with rigid or rule-based templates         \citep{renz2025simlingo,lc_llm,nvidia2025alpamayo}. 
However, reliance on these explanations is risky: recent work shows that CoT reasoning is often an unfaithful 
rationalization that does not reflect the features that causally drive
the prediction \citep{turpin2023unfaithful_cot}. Yet, \citet{barez2025cot_not_xai} report that 63\% of recent autonomous systems papers treat CoT as an \emph{interpretability} method.
In safety-critical domains, such unfaithful explanations can mask the true causes of dangerous behaviors, creating a strong need for methods that directly identify the input factors responsible for catastrophic decisions.

\begin{figure}
    \centering
    \includegraphics[width=\columnwidth]{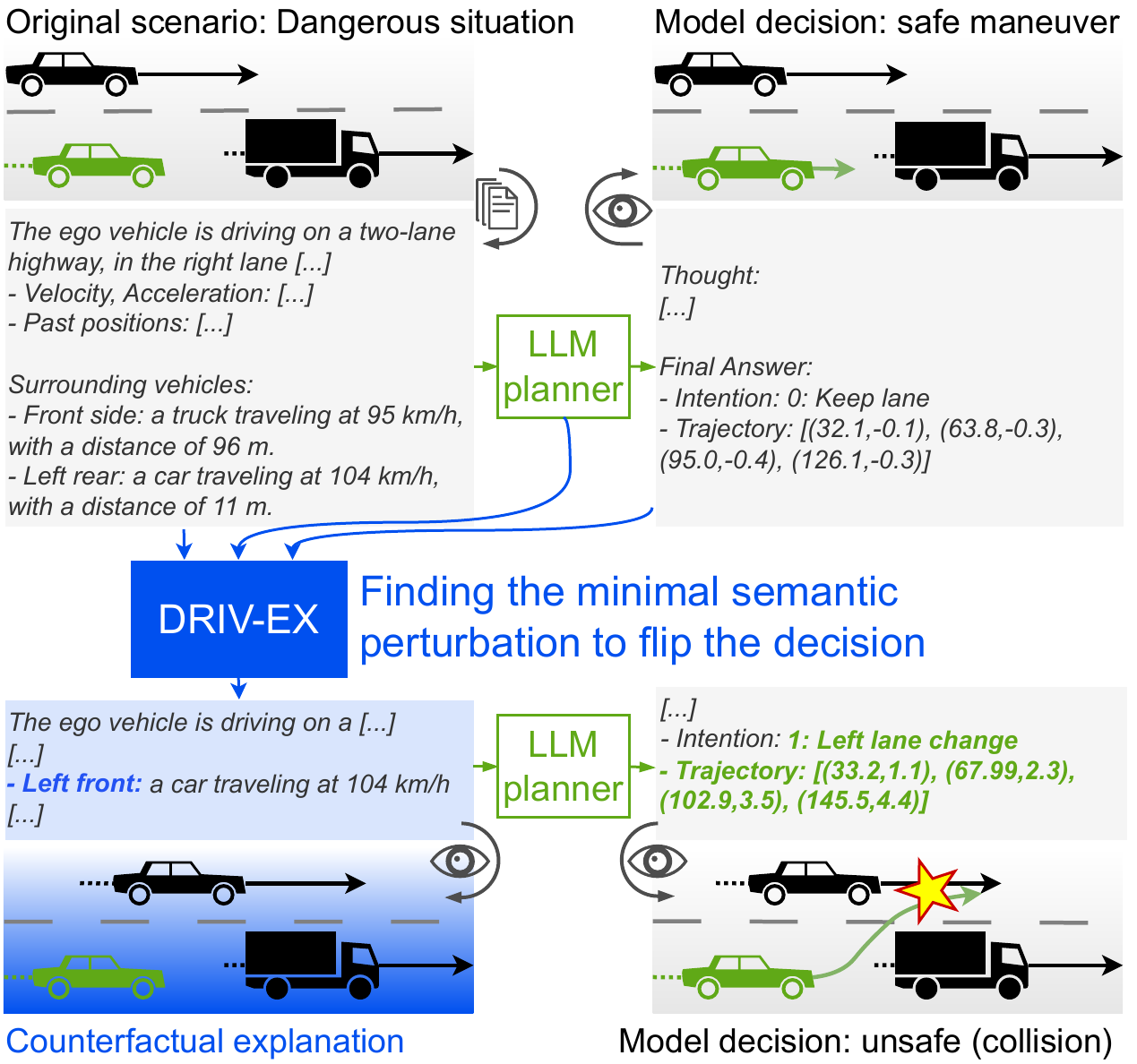}
    \caption{\textbf{Overview of DRIV-EX counterfactual generation.} The LLM acts as the planner for the ego vehicle (in green). Given an initial driving scenario where the planner behaves safely (top row), our method automatically identifies a minimal semantic perturbation to the scene description (such as slightly altering the position or speed of surrounding vehicles) that forces the model into a dangerous failure mode (bottom row). By uncovering these decision boundaries, DRIV-EX exposes latent biases and evaluates the robustness of driving agents against critical edge cases.}
    \label{fig:teaser}
\end{figure}

Counterfactual explanations offer a grounded alternative to narrative-based insights. By identifying the minimal change to an input that would lead the model to make a different decision, counterfactuals explicitly isolate the causal features of a behavior. 
In the context of LLM-based driving, this involves finding the smallest modification to a scene description that turns a safe maneuver into a collision (or vice versa). As illustrated in \autoref{fig:teaser}, this provides a concrete mechanism to audit decision boundaries and expose latent safety risks.

Technically, generating these textual counterfactuals requires solving a difficult discrete optimization problem: finding a sequence of tokens that is semantically close to the original description yet validly flips the driving decision. This presents a dilemma between two standard approaches.
\emph{Gradient-based methods} (often used in `soft' prompt optimization) relax discrete text into continuous embeddings to efficiently search for decision boundaries, but they generally fail to recover coherent text when projected back to the discrete vocabulary \citep{kumar2022mucola,wen2023pez}.
Conversely, \emph{discrete search methods} that rely on the model's native autoregressive sampling ensure fluency but lack the gradient guidance needed to efficiently cross the decision boundary \citep{liu2023bolt,pynadath2025dab}.
Lastly, satisfying the \emph{minimality} constraint is particularly challenging: without robust regularization, optimization algorithms tend to drift far from the original input, hallucinating entirely new objects or contexts rather than isolating the subtle feature shifts that explain the model's behavior.

In this work, we introduce \ours (DRIVing EXplanations), an algorithm designed to resolve this tension.
Our key idea is to decouple the \emph{search} for the counterfactual from its \emph{generation}.
\ours treats optimized continuous embeddings from gradient-based methods not as the final output, but as a \emph{semantic guide} for a controlled decoding framework.
Practically, our algorithm performs gradient-based updates on the unconstrained prompt embeddings to identify the direction required to flip the decision.  Then, it uses these embeddings to bias an autoregressive re-generation of the input.
This allows us to steer the fluent generation capabilities of the LLM toward the critical decision boundaries identified by the gradients, effectively combining the search power of continuous optimization with the coherence of autoregressive generation.

We evaluate \ours through quantitative experiments on LC-LLM \citep{lc_llm}, a recent driving LLM planner, using a textual transcription of the highD dataset \citep{highd}, a real-world highway driving benchmark. We show that \ours generates valid counterfactuals more reliably than baselines \citep{pynadath2025dab,wen2023pez,univ_attacks}. We further use \ours to reveal biases that push models toward unsafe decisions. Finally, we demonstrate that these insights can be used to mitigate biases, improving the safety and robustness of driving LLMs.

\section{Related Work}
\label{sec:related_work}

\subsection{LLMs as driving agents}
LLMs are becoming a dominant approach for trajectory planning in autonomous driving. LLMs were first proven able to perform planning, translated into a next token prediction task, relying either on text tokens \citep{gpt_driver, lc_llm, trajectory_llm}, or on additional tokens accounting for discretized motions, trajectories or actions \citep{motion_lm, smart}. 
Recently, Multimodal LLMs (MLLMs), that also process images and videos, have been increasingly used for end-to-end driving. 
Notably, their performance is often boosted by a Chain-of-Thought mechanism, where text tokens are predicted to `reason' on the scene, right before planning. These reasoning traces are often structured with rigid or rule-based templates. Upon release, some of these methods were state of the art on major driving benchmarks, such as EMMA \citep{hwang2025emma} on nuScenes \citep{nuscenes}, Poutine \citep{rowe2025poutine} on WOD-E2E \citep{waymo_dataset} 
and SimLingo \citep{renz2025simlingo} on Bench2Drive \citep{Bench2Drive}.

\subsection{Counterfactuals for driving (M)LLMs}

Counterfactuals for (M)LLM-based driving have emerged in distinct paradigms. First, simulation-based approaches embed counterfactual reasoning within the model's decision process. For instance, in \citet{cf_vla}, the model iteratively evaluates counterfactual futures to revise its plans. 
Another approach uses counterfactuals as additional supervision for training, with data augmentation \citep{cf_responsibility, OmniDrive}. This enables models to capture causal sensitivities without performing explicit counterfactual inference at test time. 
In contrast, our work uses a classical formulation of counterfactuals as minimal input perturbations that alter a model’s decision, for offline auditing. 

\subsection{Explainability methods for NLP models}
Explainability methods for NLP models are commonly divided into local and global approaches \citep{zhao2024xai_nlp_survey}.
Local methods explain individual predictions, typically via attribution techniques \citep{li20216visualizing,wu2020perturbed,mohebbi2021exploring,enguehard2023sequential_integrated_gradients}.
Global approaches aim to uncover model-wide behaviors across inputs 
\citep{bricken2023towards,conmy2023mechanistic}, but are less suited to explaining specific decisions.

\smallskip\noindent\textbf{Counterfactual explanations} are a form of local interpretability, identifying minimal changes to an input that would alter a model’s prediction, thereby exposing causal decision factors \citep{wachter2017counterfactual}. In NLP, they have been studied for discriminative tasks such as sentiment analysis and toxicity detection \citep{wu2021polyjuice,nguyen2024llms}. Typical approaches generate counterfactuals by substituting, masking, or infilling tokens while preserving fluency and meaning \citep{wu2021polyjuice}. Most NLP counterfactual methods follow a two-stage pipeline: (i) identifying influential input tokens, and (ii) editing these tokens to induce a label change \citep{ross2021mice,fern2021closs,treviso2023crest}. For example, MiCE \citep{ross2021mice} selects tokens via gradient-based attribution and enforces minimality through constrained search, while CREST \citep{treviso2023crest} uses sparse rationales to relax strict edit minimality. TIGTEC \citep{bhan2023tigtec} similarly performs sequential masked token replacement guided by local importance.
Despite operating in discrete text space, these methods are not directly applicable to our setting as they are designed for classifiers.

\smallskip\noindent\textbf{Prompt optimization} methods search for discrete or continuous input prompts that induce desired behaviors from pretrained language models. Soft prompt tuning introduces continuous embeddings optimized via gradients without updating model parameters \citep{lester2021soft,li2021prefix}. Discrete approaches such as AutoPrompt \citep{autoprompt} and PEZ \citep{wen2023pez}  
perform gradient-based updates in embedding space followed by projections 
to discrete tokens.

Related techniques have been explored extensively in adversarial prompting, jailbreak attacks, and red teaming, where the goal is to elicit unsafe or restricted behaviors from fixed LLMs \citep{wallace2019universal,cold_attack,univ_attacks,mo2024fight}. These methods typically prioritize objective satisfaction over semantic fidelity, often producing prompts that are unnatural or not fluent.

Our work shares the underlying optimization structure of these approaches: searching over token sequences to satisfy a target objective. However, it differs in motivation and constraints. Rather than inducing arbitrary or adversarial behaviors, we seek minimal, instance-specific edits that preserve semantic plausibility while flipping a concrete decision. This places stronger constraints on fluency, proximity, and interpretability than most prompt optimization or attack-oriented methods enforce.

\smallskip\noindent\textbf{Controlled decoding} methods steer pretrained autoregressive models at inference time so that generated sequences satisfy explicit constraints while remaining fluent \citep{picard,fudge}. Approaches include biased autoregressive sampling \citep{liu2023bolt}, reinforcement learning over decoding policies \citep{CD_from_LM}, and gradient-based sampling methods combining multiple objectives \citep{cold,kumar2022mucola,pynadath2025dab}.
Several methods explicitly operate in discrete token space using local edits or MCMC-style proposals. For example, MuCoLa \citep{kumar2022mucola} performs constrained editing via MCMC to balance constraint satisfaction and likelihood, while DAB \citep{pynadath2025dab} combines discrete Langevin proposals with biased autoregressive generation to improve exploration near constraint boundaries.

While controlled decoding is closely related to our approach, key differences remain. Controlled decoding is typically framed as a generation problem without a reference input, targeting global attributes such as sentiment or topic. In contrast, our task is explicitly counterfactual: we operate relative to a specific reference prompt and enforce semantic proximity to it. This shifts the objective from general constraint satisfaction to fine-grained, local edits that expose decision boundaries. 

\smallskip\noindent\textbf{Self-Generated Counterfactual Explanations}
for LLMs is an emerging paradigm, where LLMs are prompted to generate minimally altered inputs that could flip their own decision. This bypasses the need for external counterfactual methods but \citet{can_llms_explain_themselves_2025} and \citet{llms_dont_know_2025} show that LLMs do not generate valid counterfactuals via prompting. Moreover, this paradigm is not fully adaptable to our setting: as LLM-based planners are constrained to produce formatted outputs such as trajectories, they cannot be directly prompted to generate counterfactual inputs.

\section{Method: \ours}
\label{sec:method}

\subsection{Problem and method overview}

\paragraph{Formalization.}
Let $\mathcal{M}$ be an autoregressive language model with vocabulary $V$. Given an original input sequence $\mathbf{x}^o = (x^o_1, \dots, x^o_m) \in V^m$, the model produces the sequence $\mathbf{y}^o = (y^o_1, \dots, y^o_n) = \mathcal{M}(\mathbf{x}^o)$. We focus on a particular decision made during generation, formalized as the token emitted at the semantic decoding step $T$ (e.g., the first token encoding a planning decision in a structured output). Importantly, $T$ refers to a semantic position rather than a fixed index and may depend on the generated prefix. As observed in prior work on LLM steering and jailbreaks, influencing the first critical decision token is often sufficient to redirect the entire subsequent generation \citep{Jailbroken,CarliniNCJGKITS23,univ_attacks}
since autoregressive dynamics strongly bias the continuation. This motivates focusing the counterfactual search on enforcing the constraint at step $T$. Empirical evidence for this phenomenon is given in \autoref{sec:appendix_lc_traj_coherence}.

We seek a counterfactual explanation $\mathbf{x}^\text{cf}$ that induces a target decision $y^*_T$ at step $T$, while remaining a minimal, plausible, modification of $\mathbf{x}^o$. This yields competing objectives:  
(i) enforcing the desired \textit{decision} at step $T$, 
(ii) preserving \textit{fluency}, and (iii) semantic \textit{proximity} to the initial input. We express this as a constrained optimization problem: 
\begin{equation}
\label{eq:counterfactual_objective}
\begin{aligned}
& \underset{\mathbf{x} \in V^m}{\arg\min}
& & c(\mathbf{x}^o, \mathbf{x}) \\
& \text{s.t.}
& & y^*_T = \underset{z\in V}{\arg\max} P_{\mathcal{M}}(z \mid \mathbf{y}_{<T}, \mathbf{x}) ,
\end{aligned}
\end{equation}
where $c(\mathbf{x}^o, \mathbf{x})$ is a cost function capturing proximity and fluency. 

\begin{algorithm}[t]
\caption{\ours pseudo-code}
\label{alg:counterfactual_pez}

\KwIn{
Original input $\mathbf{x}^o = (x^o_1,\dots,x^o_m)$;
target token $y^*_T$ at step $T$;
model $\mathcal{M}$ 
}
\KwOut{Counterfactual explanation $\mathbf{x}^{\mathrm{cf}}$}

Initialize soft embeddings $\mathbf{e}$ using $\mathbf{x}^o$\;

\For{$n = 1$ \KwTo $N$}{

    \emph{$\rhd$ Projection and forward pass:} \\
    Project $\mathbf{e}$ to get ${\mathbf{x}}$ \tcp*{\autoref{eq:projection}}
    Compute decision loss $\mathcal{L}_{\text{dec}}$ \tcp*{\autoref{eq:decision_loss}}

    \emph{$\rhd$ Backward (straight-through update):} \\
    Compute gradient $\nabla_{\mathbf{x}} \mathcal{L}_{\text{dec}}$\;
    Update soft embeddings $\mathbf{e}$ \tcp*{\autoref{eq:gradient_step}}

    \emph{$\rhd$ Regularized autoregressive decoding:} \\
    Compute biases $\mathcal{B}$ and $\mathcal{B}'$ from $\mathbf{e}$ and $\mathbf{x}^o$\tcp*{Eq.~\ref{eq:voc_penalization}--\ref{eq:bias}}
    Decode candidate $x$ with biased decoding \tcp*{\autoref{eq:biased_sampling}}
    
}

\KwRet $\mathbf{x}^\text{cf}$\tcp*{Best candidate $x$}
\end{algorithm}

\paragraph{High-level method idea.}
Directly optimizing the counterfactual objective in \autoref{eq:counterfactual_objective} is challenging because the input $\mathbf{x}$ lies in a discrete space and the autoregressive decoding process of LLMs is non-differentiable. Naively relaxing the input to a continuous space enables gradient-based prompt optimization with straight-through gradients \citep{autoprompt,wen2023pez}, but discretizing the resulting embeddings often leads to incoherent or out of distribution text, violating the plausibility requirement.

To address this, we decouple the \emph{search} for the counterfactual from its \emph{generation}. We utilize the optimized continuous embeddings not as final outputs but as a \emph{semantic guide} within a controlled decoding framework \citep{liu2023bolt,pynadath2025dab}. Specifically, we perform gradient-based updates on the embeddings to identify the direction required to flip the decision. We then use these embeddings to bias an autoregressive re-generation of the input, steering the sampling toward the identified decision boundary and the original text while strictly enforcing fluency. The complete procedure is summarized in \autoref{alg:counterfactual_pez}.

\subsection{Discrete counterfactual optimization via straight-through embeddings}
\label{sec:method:optimization}

\paragraph{Continuous relaxation.}
We associate each input token with a continuous embedding and maintain a sequence of soft embeddings
$\mathbf{e} = (\mathbf{e}_1, \dots, \mathbf{e}_m) \in \mathbb{R}^{m \times d}$, initialized from the embeddings of the original input $\mathbf{x}^o$. At each iteration, the current soft embeddings are projected onto the nearest vocabulary embeddings using cosine similarity:
\begin{equation}
\mathbf{x}_i = \text{Proj}_E(\mathbf{e_i}) := \arg\max_{\mathbf{v} \in E} \frac{\mathbf{e}_i \cdot \mathbf{v}}{\|\mathbf{e}_i\|\|\mathbf{v}\|},
\label{eq:projection}
\end{equation}
yielding a discrete token sequence $\mathbf{x}$ compatible with the LLM's vocabulary.

\begin{figure}[t]
    \centering
    \includegraphics[trim={0 0 0.9cm 0},clip,width=\linewidth]{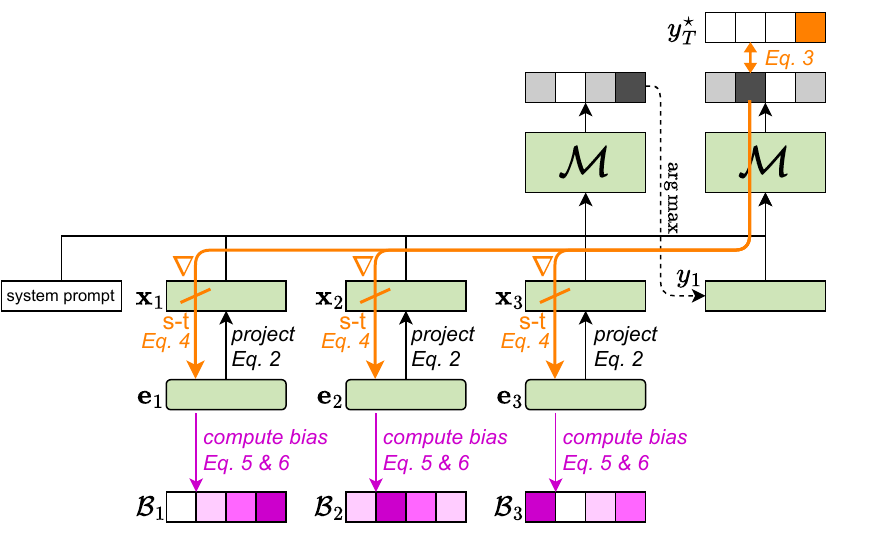}
    \caption{\textbf{Forward and backward passes, and bias computation.} In the forward pass (black), the continuous soft embeddings $\mathbf{e}$ are projected onto their nearest neighbors in the vocabulary to obtain tokens $\mathbf{x}$ (\autoref{eq:projection}). These tokens are processed by the model $\mathcal{M}$ to compute the probability of the target token $y^*_T$ (\autoref{eq:decision_loss}). 
    In the backward pass (orange), gradients ($\nabla$) are backpropagated and, using a straight-through (s-t) estimator, bypass the discrete projection to update the continuous embeddings $\mathbf{e}$ (\autoref{eq:gradient_step}). Finally, the updated embeddings are converted into vocabulary bias terms ($\mathcal{B}$, in pink) to guide subsequent regularization (following Eq.~\ref{eq:voc_penalization}--\ref{eq:bias}).}
    \label{fig:backward}
\end{figure}

\paragraph{Decision-driven gradient update.}
The projected sequence $\mathbf{x}$ is fed to the model $\mathcal{M}$ to evaluate the decision objective at step $T$:
\begin{equation}
\label{eq:decision_loss}
\mathcal{L}_{\text{dec}}(\mathbf{x}) = - \log P_{\mathcal{M}}(y^*_T \mid \mathbf{y}_{<T}, \mathbf{x}),
\end{equation}
and compute gradients with respect to the projected embeddings linked to $\mathbf{x}$. Using a straight-through estimator \citep{bengio2013straightthrough}, these gradients are applied to the continuous embeddings $\mathbf{e}$, which are updated via Adam \citep{kingma2015adam}:
\begin{equation}
\mathbf{e} \leftarrow \text{Adam}\!\left(\mathbf{e}, \nabla_{\mathbf{x}} \mathcal{L}_{\text{dec}}, \eta \right).
\label{eq:gradient_step}
\end{equation}
This process is illustrated in \autoref{fig:backward}.

This project-forward-backward-update loop constitutes the core optimization algorithm and is repeated a predefined number of $N$ iterations.

\subsection{Projection-based regularization and evaluation}
\label{sec:method:regularization}

While the gradient-based updates in \autoref{sec:method:optimization} effectively inject the decision-change signal into the continuous embeddings $\mathbf{e}$, directly projecting these embeddings onto the vocabulary often yields incoherent sequences that fail to satisfy the plausibility constraint. To bridge this gap, we employ a regularized autoregressive decoding strategy \citep{liu2023bolt,pynadath2025dab} where the optimized embeddings serve as a \emph{semantic guide} to steer a dedicated fluency model $\mathcal{F}$. 
We instantiate $\mathcal{F}$ using the same pre-trained model as $\mathcal{M}$, yet without planning-oriented adaptation, to guarantee a shared vocabulary and tokenization, enabling us to directly bias its logits using distances computed in the common embedding space.

\paragraph{Fluency regularization with biased autoregressive decoding.}
For each position $i$, we compute a distance vector $\tilde{\mathbf{b}}_{i} \in \mathbb{R}^{|\mathcal{V}|}$ measuring how far each vocabulary token $v$ is from the projected token $\text{Proj}_E(\mathbf{e_i})$ in embedding space:
\begin{equation}
\label{eq:voc_penalization}
b_{i,v} = \lVert \mathbf{v} - \text{Proj}_E(\mathbf{e_i}) \rVert_2^2.
\end{equation}

To match the scale of the model logits $\mathbf{l}_i = \text{logits}_{\mathcal{F}}(\cdot \mid x_{<i})$, we compute a normalization factor
$r_i = \lVert \mathbf{l}_i \rVert_2 / \lVert {b}_i \rVert_2$ as in \citet{pynadath2025dab}. The bias applied to each token $v$ is then
\begin{equation}
\label{eq:bias}
\mathcal{B}_{i,v} = - w . r_i . b_{i,v},
\end{equation}
where $w$ controls the strength of the bias.

\begin{figure}
    \centering
    \includegraphics[trim={1.25cm 0 0.9cm 0},clip,width=\linewidth]{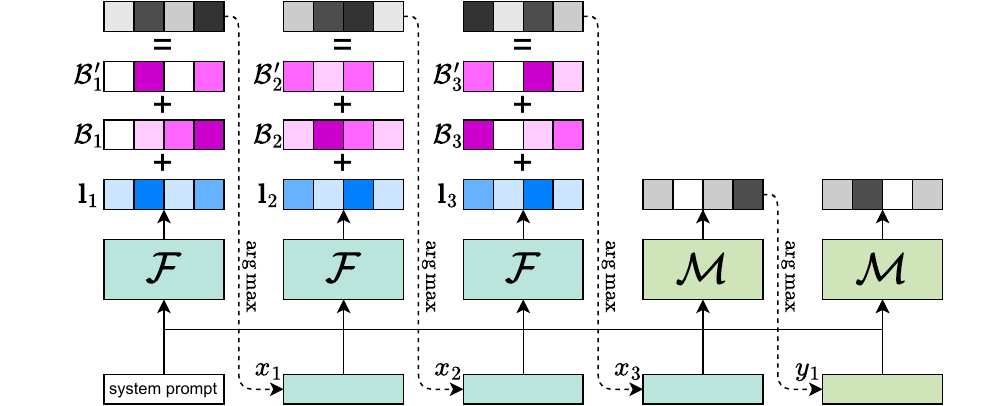}
    \caption{\textbf{Regularized autoregressive decoding.} During the regularization phase, the vocabulary bias terms $\mathcal{B}$ and $\mathcal{B}'$ (derived from optimized embeddings and $\mathbf{x}^o$) are added to the logits $\mathbf{l}$ of a fluency model $\mathcal{F}$. This combined signal biases the auto-regressive decoding, following \autoref{eq:biased_sampling}, to generate a new candidate sequence ($x_1, x_2, x_3$) that incorporates the decision-change signal while maintaining fluency and input proximity.}
    \label{fig:regularization}
\end{figure}

The bias is added to the logits of the fluency model $\mathcal{F}$ during greedy decoding, encouraging generations that are close to the optimized embeddings while preserving fluency. 
Concretely, as illustrated in \autoref{fig:regularization}, the next token is chosen with: $x_i = \arg\max_{v \in \mathcal{V}} \left( l_{i,v} + \mathcal{B}_{i,v} \right)$.

The resulting decoded sequence provides a concrete counterfactual candidate that can be evaluated against the objective in \autoref{eq:counterfactual_objective}. Importantly, this regularization does not define a separate optimization step but serves to monitor and enforce plausibility throughout the optimization trajectory.

\paragraph{Input proximity regularization.}
\label{subsec:input_prox_reg_methods}
We propose three complementary approaches to ensure semantic proximity to $\mathbf{x}^o$. 
First, `\textbf{Proj}' implies to limit embedding projections (\autoref{eq:projection}) to the subset of the top $K$ nearest neighbors of $\mathbf{x}^o$ (in cosine similarity).
Second, `\textbf{Loss}' adds $(1 - \frac{1}{m} \sum_{i=1}^{m}\cos \text{sim}\!\left( \mathbf{x}_i,\, \mathbf{x}^o_i \right))$ as a regularization term to the decision loss (\autoref{eq:decision_loss}), controlled by a weight $\lambda$. 
Finally, `\textbf{Bias}' adds a second bias during autoregressive biasing (\autoref{eq:biased_sampling}).
Using $\mathbf{x}^o_i$ instead of the current projected token $\text{Proj}_E(\mathbf{e_i})$ in \autoref{eq:voc_penalization}, we compute $\mathcal{B'}_{i,v}$ following Eq.~\eqref{eq:voc_penalization}--\eqref{eq:bias}, and the biased sampling becomes:
\begin{equation}
\label{eq:biased_sampling}
x_i = \arg\max_{v \in \mathcal{V}} \left( l_{i,v} + \mathcal{B}_{i,v} + \mathcal{B}'_{i,v} \right).
\end{equation}

\paragraph{Best candidate selection.}
Upon completion of $N$ iterations, we use a heuristic selection to derive the best counterfactual from the pool of generated candidates. We prioritize candidates that successfully trigger the decision change; specifically, we define the set of valid candidates as those satisfying 
$y^*_T = {\arg\max}_{z \in V} P_{\mathcal{M}}(z \mid \mathbf{y}_{<T}, x)$.
If this set is non-empty, we return the candidate that maximizes semantic proximity to the original input $\mathbf{x}^o$, as the autoregressive decoding is already ensuring high fluency. Conversely, if no candidate triggers the decision change,
we return the sequence that achieves the highest target probability $P_{\mathcal{M}}(y^*_T  \mid \mathbf{y}_{<T}, x)$. 

\section{Experiments}
\label{sec:expe}

We evaluate \ours on LC-LLM~\citep{lc_llm}, a LoRA-finetuned planner \citep{lora}, on a textual transcription of the highD dataset~\citep{highd}.
This dataset maps traffic annotations to Lane Changes (LC) and trajectories.

\begin{table*}[t]
  \centering
  \begin{tabular}{l l  c c c c c c}
    \hline
   &  & \textbf{Flip rate}  
   & \multicolumn{2}{c}{\textbf{Fluency}}  & \multicolumn{1}{c}{\textbf{Similarity}}  & \multicolumn{2}{c}{\textbf{Aggregation}} \\
   \cmidrule(lr){3-3} \cmidrule(lr){4-5} \cmidrule(lr){6-6} \cmidrule(lr){7-8}
   &  & 
    \textbf{$y_T^*$ has top $\uparrow$} 
   & \textbf{Min $\uparrow$} 
    & \textbf{Template $\uparrow$}  & \textbf{BERTSc $\uparrow$}  
    & \textbf{Aggreg $\uparrow$} & \textbf{Aggreg $\uparrow$} \\
  \textbf{Method}  &  \textbf{LLM} & 
  \textbf{rank (\%)} 
  & \textbf{fluency}  & \textbf{filter (\%)} & \textbf{filter (\%)}  
  & \textbf{(\%)}  & \textbf{\& col (\%)}  \\
    \hline
   \verb|DAB|  & \small{Llama3} & 100.0  & 9e-4 & 86.7 & 33.3 & 33.3 & 14.7 \\
   \multirow{2}{*}{\parbox{1.5cm}{\footnotesize \citep{pynadath2025dab}}}  &  \small{Mistral} & 91.8 & 6e-3 & 30.6 & 26.5 & 24.5 & 20.4
  \\
  &  \small{Qwen2.5}  & 98.9 & 2e-3 & 7.8 & 10.0  & 7.8 & 7.8
  \\
 \hline
   \verb|DAB| $\dagger$ & \small{Llama3} & 46.7 & 5e-3 & 100.0  & 94.7 & 41.3 & 30.7
 \\
 &  \small{Mistral} & 40.8 & 3e-2 & 98.0 & 91.8 & 38.8 & 34.7 \\
            &  \small{Qwen2.5}  & 26.7 & 7e-3 & 93.3 & 93.3 &  24.4 & 21.1
 \\
 \hline
    \verb|PEZ|  &  \small{Llama3} & 97.3 & 2e-4 & 45.3 & 81.3 & 45.3 & 37.3 \\
    \multirow{2}{*}{\parbox{1.5cm}{\footnotesize \citep{wen2023pez}}} &  \small{Mistral}  & 100.0 & 5e-4 & 42.9 & 89.8 & 42.9 & \underline{40.8} \\
            &  \small{Qwen2.5}  & 96.7 & 2e-3 & 23.3 & 44.4 & 23.3 & 17.8
 \\
    \hline
    \verb|PEZ| $\dagger$ &  \small{Llama3} & 97.3 & 2e-4 & 49.3 & 86.7 & 49.3 & 37.3 \\
            &  \small{Mistral} &  100.0 & 1e-3 & 57.1 & 95.9 & \underline{57.1} & \underline{40.8} \\
            & \small{Qwen2.5}  & 94.4 & 3e-3 & 30.0 & 58.9 & \underline{30.0} & \textbf{23.3} \\
    \hline 
\verb |GCG| 
& \small{Llama3}  & 98.7 & 2e-4 & 60.0 & 100.0 & \underline{58.7} & \underline{44.0} \\
 \multirow{2}{*}{\parbox{1.5cm}{\footnotesize \citep{univ_attacks}}} & \small{Mistral} & 98.0 & 1e-4 & 20.4 & 100.0 & 20.4 & 18.4  \\
&  \small{Qwen2.5} & 88.9 & 5e-4 & 13.3 & 98.9 & 13.3 & 8.9
 \\
    \hline
 \rowcolor{nicegreen!20!white}    \verb|DRIV-EX| 
 &  \small{Llama3} & 64.0 & 1e-3 & 88.0 & 96.0 & \textbf{61.3} & \textbf{56.0} \\
  \rowcolor{nicegreen!20!white}  (ours)  &  \small{Mistral} &  83.7 & 4e-3 & 79.6 & 95.9 & \textbf{69.4} & \textbf{61.2}  \\
 \rowcolor{nicegreen!20!white}      &  \small{Qwen2.5}  &  58.9 & 3e-3 & 47.8 & 73.3 & \textbf{34.4} & \underline{22.2} \\
    \hline
  \end{tabular}
  \caption{\textbf{Main counterfactual explanation results} on safety-critical textual highD subsets, where we steer LLMs from a safe decision to a collision. We give results for comparable compute budgets: N=15 iterations for DAB/DRIV-EX, N=150 for PEZ, except for the compute-heavy GCG for which N=6 leads to 5 times more compute. `$\dagger$' denotes task-adapted baselines (vs.\ off-the-shelf), `BERTSc' stands for `BERTScore'.}
  \label{tab:algo_comparison}
\end{table*}

\subsection{Experimental Protocol}
\label{sec:expe:protocol}

We focus on probing the brittleness of the model's safety predictions.
From the validation set, we select `\emph{dangerous}' scenarios, defined as those where the ground-truth trajectory implies a potential collision (impact within 4s at constant velocity). We further focus on scenarios where a given model predicts a `\emph{safe}' maneuver.
To assess robustness, we investigate whether minimal semantic perturbations to these scenes can force the model into a catastrophic failure.
We set the target $y^*_T$ to the ground-truth lane change token, and employ \ours and baselines to flip the model's decision from `\emph{safe}' to `\emph{dangerous}'.
For all methods, updates are restricted to tokens related to words that can change in the driving template (e.g., vehicle type, speed, position). These tokens are shown in \autoref{sec:appendix_ft_bias}, \autoref{fig:unbiased_sample} and subset statistics are given in~\autoref{tab:crash_subset}. This setup, designed for offline model auditing, allows us to (1) benchmark \ours against existing baselines (\autoref{sec:expe:baseline_ablation}), (2) uncover model biases that trigger unsafe decisions (\autoref{sec:expe:eval_robustness}), and (3) demonstrate how these insights can help mitigate bias and improve safety (\autoref{sec:expe:improving}).

\subsection{Comparison with baselines and ablations}
\label{sec:expe:baseline_ablation}

We compare \ours against DAB \citep{pynadath2025dab}, PEZ \citep{wen2023pez}, GCG \citep{univ_attacks} and their task-adapted versions ($\dagger$). All methods are used on the LC-LLM planner, implemented with instructed versions of Llama3-8B \citep{llama3herdmodels}, Mistral-7B \citep{mistral} and Qwen2.5-7B \citep{qwen_2_5}, as detailed in \autoref{sec:appendix_ft_normal}. Baselines description, adaptation and hyperparameter search are detailed in~\autoref{sec:appendix_hyperparam}.

\paragraph{Evaluation Metrics.}
\label{sec:expe:eval_metrics}
We track metrics that relate to the counterfactual explanation constraints, given in \autoref{eq:counterfactual_objective}. We describe them below; further details on metrics and thresholds are given in \autoref{subsec:appendix_bertscore_threshold}.

$\bullet$ \textbf{Decision (flip rate):} Ratio of candidates that lead to the target $y^*_T$ being top-ranked at step $T$. 

$\bullet$ \textbf{Similarity and Fluency:} BERTScore~\citep{bert_score} is a measure of semantic proximity between sequence pairs. Only counterfactuals with a BERTScore $\ge 0.95$ with the initial input are kept. In \textbf{BERTScore filter}, we give the ratio of such counterfactuals.
\textbf{Template Fitness filter} quantifies the ratio of counterfactuals that perfectly respect the driving template.
\textbf{Min Fluency} is the minimum conditional token probability (cf \autoref{sec:appendix_metrics}).

$\bullet$ \textbf{Aggregated Scores (Success Rate):} These are our primary metrics that we optimize. \textbf{Aggreg} measures the number of counterfactual `Successes', where a success satisfies three strict conditions: the decision is flipped, the template is perfectly valid and semantic similarity is high (BERTScore $\geq 0.95$). 
We also report \textbf{Aggreg \& Col}, the subset of successes that lead to a collision. It quantifies the method's ability to expose dangerous behaviors.

\paragraph{Quantitative results.}
Generating counterfactuals requires balancing decision flipping with similarity and fluency. As shown in \autoref{tab:algo_comparison}, off-the-shelf DAB achieves a high flip rate (100\% for Llama3) but fails on similarity (BERTScore filter of 33.3\%) due to a highly exploratory proposal step. Adding proximity regularization (DAB$\dagger$) improves similarity but drastically degrades the decision flip rate (46.7\%).
Conversely, PEZ (and its adapted version) maintains high flip rate and similarity to the original scene but lacks fluency, resulting in a lower `Aggregated Score' (45.3\% vs.\ 61.3\% for \ours on Llama3).
GCG is a strong baseline on Llama3, reaching the second best `Aggregated Score' of 58.7\%. However, its performance is weak on other LLMs, particularly so on Mistral where it has the lowest `Aggregated Score' (20.4\%), due to poor token replacement proposals. This variation suggests that GCG is sensitive to the embedding space geometry of LLMs.
\ours outperforms all baselines in `Aggregated Score', for all LLMs, by effectively using the fluency model $\mathcal{F}$ to guide generation. This ensures counterfactuals that are valid, fluent and semantically close to the original scene (96.0\% BERTScore filter on Llama3).

\paragraph{Scaling performance.}

\begin{figure}[h]
  \includegraphics[width=\linewidth]{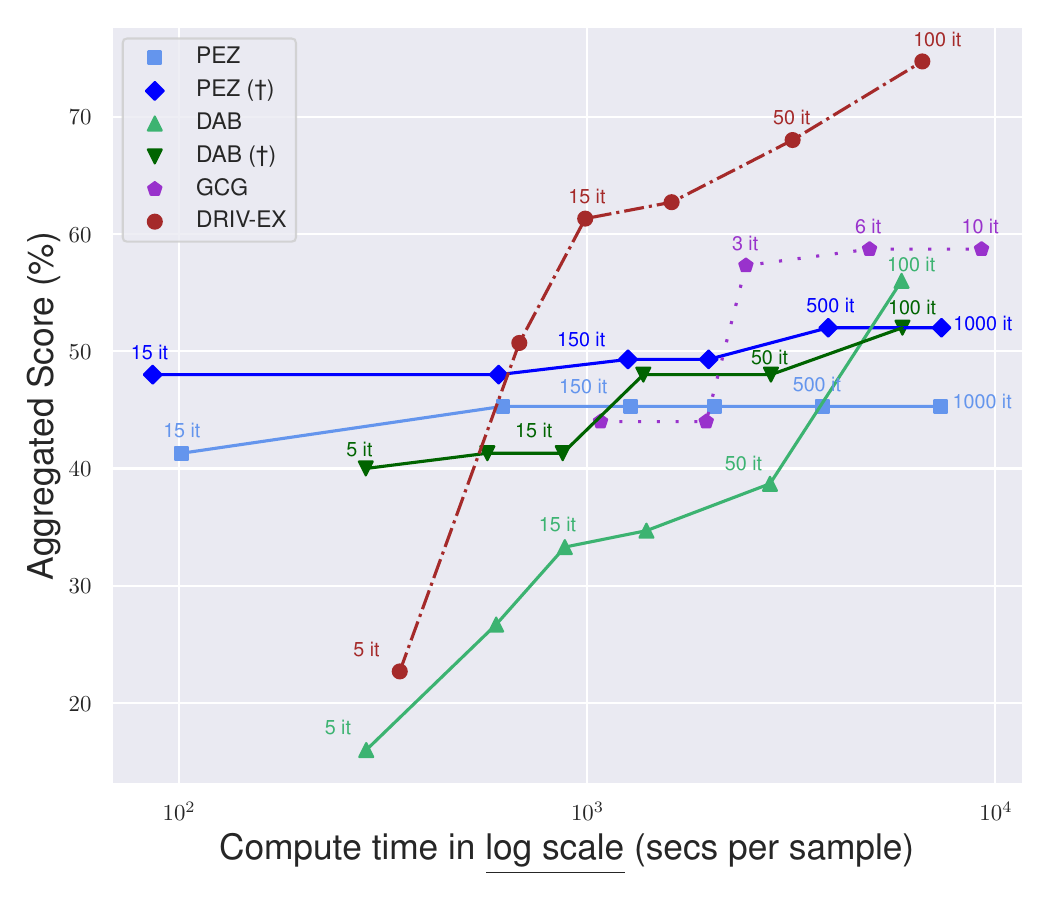}
  \caption{ \textbf{Performance scaling with respect to compute}, evaluated for all methods on Llama3 with A100 GPUs. `it' gives the number of iterations above runs.
  } 
\label{fig:compute_vs_perf}
\end{figure}

We show in \autoref{fig:compute_vs_perf} how performance in `Aggregated Score' scales with compute for \ours versus baselines. All runs are done on 40G A100 GPUs for fair comparison. 
Though PEZ and PEZ$\dagger$ are reaching competitive results for a small compute budget, their performances hardly scale with more iterations. This suggests that their best candidates are identified in early iterations, before the unregularized optimization damages the fluency.
GCG has competitive results on Llama3 but its compute cost is high due to the systematic evaluation of sampled candidates, despite batched evaluations in our implementation ($\text{batch size}=16$). Moreover, its performance barely scales beyond 3 iterations. In contrast, DAB is the only baseline that exhibits similar scaling to DRIV-EX, but with lower performance, while DAB$\dagger$ has stagnating results. 
Finally, DRIV-EX is the method that scales the most with compute, achieving by its $15^{\text{th}}$ iteration a higher `Aggregated Score' than all other baselines.

\paragraph{Ablation on proximity regularization.}
\autoref{tab:reg_ablation} evaluates the components ensuring semantic proximity to the original input, introduced in \autoref{sec:method:regularization}.
Both `Bias' and `Loss' terms improve the `Aggregated Score' (+2.7\% and +6.7\% respectively), and they are used in our adapted ($\dagger$) versions of PEZ and DAB. While `Proj' alone reduces success, combining it with `Bias' and `Loss' yields the best proximity and performance (+8.0\% in `Aggreg \& Col'), confirming the synergy of our regularizations.

\begin{table}[t]
\centering
\resizebox{\columnwidth}{!}{%
\begin{tabular}{@{}l ccc@{}}
    \hline
 \textbf{Proximity} & \textbf{BERTSc$\uparrow$} & \textbf{Aggreg$\uparrow$} & \textbf{Aggreg$\uparrow$} \\
 \textbf{regularization} & \textbf{filter}(\%) & \textbf{(\%)} & \textbf{\&Col(\%)} \\
    \hline
$\mathcal{B}$ (base, no reg)  & 82.7 & 52.0 & 48.0 \\
$\mathcal{B}$ + B (Bias) & 93.3 & 54.7 & 45.3 \\
$\mathcal{B}$ + P (Proj) & 74.7 & 50.7 & 46.7 \\
$\mathcal{B}$ + L (Loss) & 90.7 & 58.7 & 50.7 \\
 \rowcolor{nicegreen!20!white} $\mathcal{B}$ + B + P + L &  96.0 &  61.3 & 56.0\\
    \hline
\end{tabular}
}
  \caption{\textbf{Ablation of \ours proximity regularization strategies.} We compare the impact of `Bias' (B), `Proj' (P), and `Loss' (L) terms for Llama3. }
\label{tab:reg_ablation}
\end{table}

\paragraph{Qualitative result.} \autoref{fig:qual_result_cf} displays a counterfactual explanation generated with \ours. We observe that by adjusting the ego vehicle's longitudinal speed (`$v_x$') and reducing a surrounding vehicle's speed (from 98.21 km/h to 93.21 km/h), \ours successfully identifies a valid, minimal perturbation that forces the planner into a collision.

\subsection{Evaluating driving LLMs' robustness}
\label{sec:expe:eval_robustness}

\subsubsection{Retrieving injected biases}
\label{subsection:injected_bias}

To validate \ours, we first assess its ability to recover known, artificial biases injected into the training data of the driving planner. We finetune two instances of the Llama3-based planner with the following `shortcuts': (1) \textbf{`Vehicle bias'}: the model is trained to associate the ego-vehicle type (car vs. truck) with specific lane-change decisions; (2) \textbf{`Digit bias'}: the model is trained to rely on the last digit of surrounding vehicles' speeds to determine its maneuver. More details in \autoref{sec:appendix_ft_bias}.

These biased models perform similarly to unbiased models on data matching their training biases. However, their safety significantly degrades when tested on unbiased or `inversely-biased' samples (details in \autoref{tab:biased_templates}). This highlights a critical risk: LLM planners can appear competent while actually relying on non-safety-critical `surface' features.

\begin{table}[t]
\centering
\resizebox{\columnwidth}{!}{%
\begin{tabular}{@{}lclcc@{}}
    \hline
 & \textbf{\# of} & \textbf{Change}
& \multicolumn{2}{c}{\textbf{Token change (\%)}} \\
\cmidrule(lr){4-5}  
\textbf{Bias} & \textbf{data} &  \textbf{type}
& \textbf{Biased} & \textbf{Unbiased} \\
    \hline
\multirow{2}{*}{Vehicle}  & \multirow{2}{*}{19} & any & 89.5    & 8.0 \\ 
  &  & revealing & 84.2   & / \\  
 \midrule
\multirow{2}{*}{Digit}    & \multirow{2}{*}{135} & any & 52.9 & 5.3  \\  
  &  & revealing & 34.0   & / \\  
    \hline
\end{tabular}
}
  \caption{\textbf{Token change (\%) between $\mathbf{x}^\text{cf}$ and $\mathbf{x}^o$ when using DRIV-EX}. Results are split between biased and unbiased token positions. For biased tokens, `revealing' indicates cases where token values change accordingly to the injected bias, `any' for all cases.}
\label{tab:biased_tokens_change}
\end{table}

\begin{figure*}[t]
  \includegraphics[width=\linewidth]{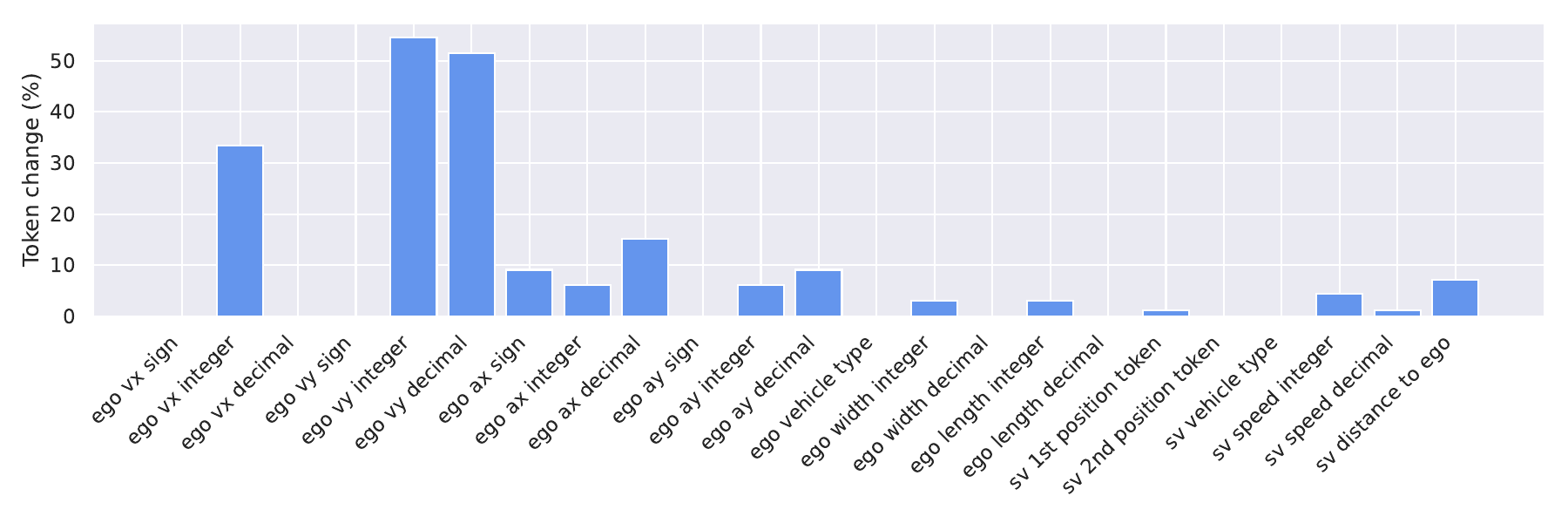}
  \vspace{-1cm}
  \caption{\textbf{Histogram of number of token changes (\%) across input position for successful counterfactuals} (Llama3) that flip a `Keep lane' decision to a collision-inducing `Right lane change' ($n$=33 samples). Peaks indicate tokens most critical for the decision flip. `sv': `surrounding vehicle', `vx/vy/ax/ay' is for velocity and acceleration.}
\label{fig:hist_keep_to_right}
\end{figure*}

To evaluate whether \ours can successfully find latent `shortcuts', we apply it to biased planners in safety-critical scenarios, and track the modification frequency  of `biased' versus `unbiased' token positions. 
We focus on samples where biased token positions have unbiased values in the initial scene, while the counterfactual is a success.
\autoref{tab:biased_tokens_change} confirms that biased token positions are significantly more susceptible to modification across both experiments. In the vehicle bias case, \ours modifies the shortcut token in 89.5\% of cases, compared to 8.0\% for unbiased positions. In 84.2\% of cases, changes are `bias-revealing,' meaning \ours directly unveils the injected shortcut, such as changing a `car' to a `truck' to induce the decision change.
Similarly, in the digit bias case, \ours modifies the subtle trailing digits in 52.9\% of cases, with 34\% of cases being bias-revealing. 
Qualitative results are shown in \autoref{fig:vehicle_cf} and \autoref{fig:digit_cf}.
These findings confirm that \ours effectively isolates specific, even non-obvious, input features that a model relies on to make driving decisions.

\subsubsection{Identifying unknown biases}
\label{subsection:unknown_bias}

We now aim at finding biases in planners trained on the original textual highD dataset. We apply \ours to the Llama3-based planner, and focus on two situations: `Keep lane' was initially inferred and we target either `Right lane change' or `Left lane change', as it covers the majority of valid counterfactuals leading to crashes (78\% and 19\%).

Without prior knowledge of existing biases, we track the percentage of token switches across all positions for counterfactuals that are both a success and lead to a collision. For `Keep lane' $\rightarrow$ `Right lane change', the histogram of token switch frequency in \autoref{fig:hist_keep_to_right} reveals that lateral velocity tokens (`$v_y$') are highly sensitive, with values switching in more than 50\% of successful cases. Manual inspection of the counterfactuals reveals that higher negative $v_y$ values strongly bias the LLM toward selecting a `Right lane change', even when it leads to a collision. Specifically, while original $v_y$ values range from $[-0.43, 0.0]$, \ours identifies that shifting these to $[-4.0, -0.47]$ flips the decision.
This shows that the model has learned to associate high negative lateral velocity with a rightward maneuver, prioritizing this shortcut over safety.
Additional results in \autoref{subsec:appendix_unknown_biases} show that Llama3 is also biased towards a left lane change when the ego car's longitudinal velocity exceeds a threshold.

Finally, an analysis on Mistral and Qwen2.5 (cf \autoref{subsec:appendix_unknown_biases}) suggests that, 
although to a lesser extent than Llama3, they also are sensitive to the ego car's velocity. Besides, they appear to be sensitive to tokens related to surrounding vehicles, in particular Qwen2.5 when it comes to other cars' speeds and distances to ego.  
These results confirm that \ours can effectively identify features that lead to unsafe behaviors in driving planners.

\subsection{Mitigation for Enhanced Robustness}
\label{sec:expe:improving}

The results in \autoref{sec:expe:eval_robustness} suggest that LLM-based planners over-rely on lateral velocity ($v_y$) shortcuts. To test if removing these biased features improves safety, we finetune Llama3-8B on three variations of the textual highD template:
$\boldsymbol{i.}$ \textbf{Baseline} (None): Includes all velocity and acceleration information ($v_x, v_y, a_x, a_y$).
$\boldsymbol{ii.}$ \textbf{Lateral-Agnostic} (no $v_y$, $a_y$): Removes lateral kinematics to eliminate the identified $v_y$ bias.
$\boldsymbol{iii.}$ \textbf{Kinematic-Free} (no $v$, $a$): Removes velocity and acceleration data, forcing reliance on spatial positioning.

\begin{table}[t]
\centering
\resizebox{\columnwidth}{!}{%
\begin{tabular}{@{}l cc cc cc@{}}
    \hline
 \textbf{Text} &
\multicolumn{2}{c}{\textbf{Classif \small{(F1$\uparrow$)}}} &
\multicolumn{2}{c}{\textbf{Traj \small{(RMSE$\downarrow$)}}}  & \textbf{\#of} & \textbf{\%of} \\
\cmidrule(lr){2-3} \cmidrule(lr){4-5}
\textbf{modif}
& \textbf{macro} & \textbf{micro} & \textbf{lon} & \textbf{lat} & \textbf{col} & \textbf{col} \\
    \hline
None & 0.93 &  0.93  &  0.68 & 0.40 & 753 & 3.1 \\  
no $v_y,a_y$  & 0.92 &  0.92  &  0.48 & 0.51 & 736 & 3.1 \\  
no $v,a$ & 0.90 &  0.90  &  1.33 & 0.55 & 631 & 2.6 \\  
    \hline
\end{tabular}
}
  \caption{\textbf{Driving performance across debiased templates} (Llama3) on highD val. We report lane change classification scores, trajectory error, number and ratio of collisions (`col'). Removing kinematic features slightly increases trajectory error but improves safety.}
\label{tab:debiasing_templates}
\end{table}

We evaluate the resulting planners on the full textual highD validation set as shown in \autoref{tab:debiasing_templates}.
We find that removing specific kinematic features enhances planner robustness. The `Lateral-Agnostic' model maintains similar F1 and RMSE scores, while reducing collisions. On the other hand, the `Kinematic-Free' model drastically reduces collisions but degrades trajectory prediction accuracy, highlighting a trade-off between predictive performance and safety. 
While our work focuses on bias detection, this intervention study shows that \ours identifies features that can guide robustness improvements.
Finer mitigation strategies, such as counterfactual data augmentation, could also be explored to leverage DRIV-EX generations.

\section{Conclusion}
\label{sec:conclusion}

In this work, we introduce \ours, a framework that explains driving LLM decisions by generating minimal, fluent counterfactuals. By decoupling gradient-based search from biased autoregressive generation, \ours successfully bridges the gap between the precise control of continuous optimization and the coherence of discrete text generation. Our experiments on the LC-LLM planner demonstrate that \ours outperforms baselines, effectively exposing latent biases, such as over-reliance on velocity, and enabling targeted safety improvements. 
Additionally, a refinement procedure of counterfactuals is proposed in Appendix \autoref{subsec:appendix_refinement}.

\section*{Limitations}
\label{sec:limitations}

While \ours effectively identifies decision boundaries, it presents specific limitations.
First, the method currently operates solely on textual inputs. Extending \ours to Vision-Language Models (VLMs) to directly handle image-based counterfactuals is a critical avenue for future research.
Second, DRIV-EX can only be applied to white-box models.
Third, the optimization assumes a fixed sequence length; the method cannot currently generate counterfactuals that require inserting or deleting tokens relative to the original input.
Fourth, our evaluation is limited to the textual highD dataset and a specific rigid template. While the framework is generic, broader validation across diverse NLP tasks and free-form text remains to be done.
Fifth, by combining gradient-based search with autoregressive sampling, \ours inherits computational costs, making it slower per-sample than pure gradient methods.
Additionally, as a local interpretability method, aggregating individual counterfactuals into global insights is challenging. The aggregation strategies proposed in \autoref{sec:expe:eval_robustness} rely on the structured nature of the textual highD template and may not generalize to unstructured free text.
Finally, our method is intended for offline model auditing, not online deployment.  

\section*{Ethical Considerations}
\label{sec:ethics}

The primary ethical implication of this work lies in its potential for dual use. \ours automates the discovery of input perturbations that induce dangerous driving behaviors (e.g., collisions). While our intent is to audit and improve safety, this capability effectively generates corner-cases that could theoretically be used to attack autonomous systems. However, we maintain that uncovering these vulnerabilities in a controlled setting is a prerequisite for building robust defenses.
Additionally, there is a risk of over-reliance on the generated explanations. Counterfactuals are approximations and may not be exhaustive; if the explanations drift from the model's true decision mechanics or miss important ones, it could lead developers to incorrect conclusions about safety. Therefore, these insights should be treated as diagnostic tools rather than absolute guarantees of model logic.

\section*{Acknowledgements}
\label{sec:acknowledgements}

We thank levelxdata for granting us access to the highD dataset \citep{highd}.

This work was performed using HPC resources from GENCI–IDRIS (Grant 2025-AD011014446R2) and with the support of
the ANR MultiTrans (ANR-21-CE23-0032).

\bibliography{biblio}


\appendix


\vspace{22mm} 

\section*{Table of contents}
\startcontents
\printcontents{ }{1}{}

\vspace{8mm} 

\section{Appendix: Empirical evidence that targeting a single token is sufficient} 
\label{sec:appendix_lc_traj_coherence}

Our single target token design relies on the `autoregressive inertia' phenomenon, which is documented in the LLM jailbreak and steering literature. As summarized in \citet{QiPL0RBM025}, the effectiveness of manipulating a very limited token window to steer the subsequent generation is validated by several established attack methods that use, like our work, a shallow surrogate objective. Works by \citet{Jailbroken} and \citet{VegaCX024} show with \textit{Prefilling Attacks} that forcing a short prefix (e.g., "Sure, I can help") can hijack the downstream generation. 
Recently, \citet{AndriushchenkoC25} obtain a 100\% attack success rate on 10 leading LLMs by performing a random search for an adversarial suffix, guided by the log-probability of a single target token (e.g., ``Sure''). While the aforementioned studies use this vulnerability for adversarial means, our work repurposes this property for explainability.

We aim to verify empirically the `autoregressive inertia' phenomenon in our setting, by verifying that targeting the \textit{first} key token is sufficient to alter consistently the rest of the LLM generation. 
To do so, we propose to monitor the semantic coherence of LLM generations beyond step $T$, when given counterfactual input prompts.

\begin{figure}[h]
  \includegraphics[width=\linewidth]{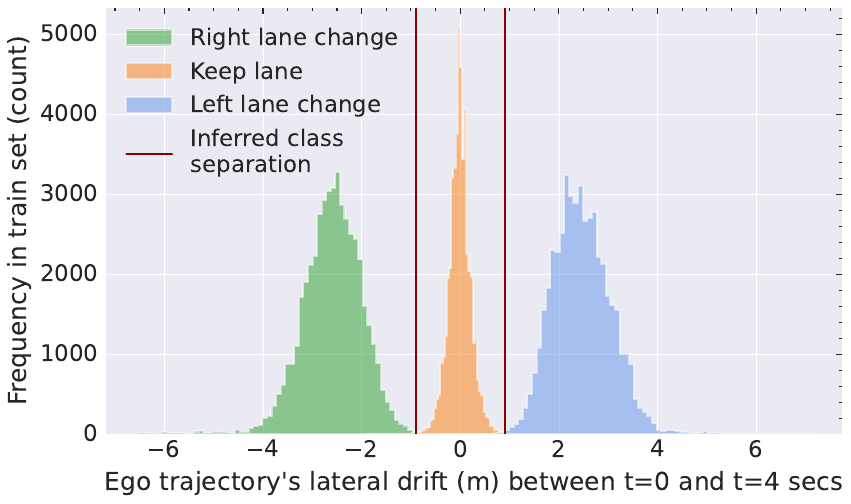}
  \caption {\textbf{Lateral drift of ground truth trajectories} per lane change class. In the coordinate system of the text templates, positive coordinate values correspond to left drifts with respect to the initial ego state, while negative values correspond to right drifts. }
\label{fig:lateral_drift}
\end{figure}

Note that the output template of textual highD first recaps notable input features before giving its inferred lane change class and planned trajectory under the form of a list of 4 coordinate tuples, one per second for the next 4 seconds. 
We propose to monitor if the lane change class is consistent with the planned trajectories, in generated outputs. To infer what would be consistent (lane change, trajectory) pairs, we first analyze the distribution of lateral trajectory drift, per lane change class, in the ground truth train set of textual highD. As shown in \autoref{fig:lateral_drift}, the samples with `Right lane change' class are always linked to a lateral trajectory drift under -0.905 m, samples with `Left lane change' class have lateral trajectory drifts above 0.905 and `Keep lane' samples have drift levels contained in [-0.905, 0.905]. The inferred class separation of 0.905 fits the data distribution and seems to correspond to $\frac{(\text{average lane width} - \text{average ego vehicle width})}{2}$, as the average lane width and average ego vehicle width are respectively of 3.75m and 1.94m in the dataset. This suggests that the data is labeled with `Left lane change' or `Right lane change' as soon as the ego vehicle has a trajectory that reaches the separation line of a neighboring lane. 

Using this inferred class separation threshold, we monitor the ratio of LLM generated outputs that lead to consistent (lane change, trajectory) pairs, when fed ground truth versus counterfactual examples as input. Results in \autoref{tab:lc_class_traj_coherence} show highly consistent outputs: for Llama3, the consistency between the flipped lane change token and the subsequently generated trajectory is 100\% on counterfactual samples, and remains above 90\% for other models.
These results support our focus on the \textit{first} key token to steer the subsequent LLM generation.

\begin{table}[h]
\centering
\begin{tabular}{ll c}
    \hline
\textbf{LLM}
& \textbf{Data}
& \textbf{Consistence (\%)}  \\
    \hline
Llama3  & GT val set & 99.8 \\
 & CF samples & 100.0 \\
Mistral  & GT val set & 99.7 \\
 & CF samples & 91.8 \\
Qwen2.5  & GT val set & 99.9 \\
 & CF samples & 90.0 \\
    \hline
\end{tabular}
  \caption{\textbf{(Lane change, trajectory) pairs consistence}, for LLM generated outputs, on ground truth (`GT') versus counterfactual (`CF') samples given by DRIV-EX.}
\label{tab:lc_class_traj_coherence}
\end{table}

\section{Appendix: Experimental details}
\label{sec:appendix_expe_detail}

\subsection{Metric details}
\label{sec:appendix_metrics}

\paragraph{Template fitness filter.} 

To compute this metric, we extract from the final decoded token sequences (under string format) whether each learnable parts of the text sample still corresponds to the template's expected type. In particular, we verify if vehicle types are comprised in \{`car', `truck'\} exclusively, if surrounding vehicles' positions are comprised within the 8 possibles candidates in the template (`Front left', `Front right', etc), and if speeds, accelerations, widths, lengths and distances are numeric data within coherent ranges, following the highD train dataset's upper and lower numeric bounds. We monitor the ratio of counterfactuals where all items correspond to the template. Penalizing counterfactuals that do not fall within the training set's numeric bounds is a way to ensure high degrees of physical plausibility for counterfactuals, despite not being an explicit guarantee.

\paragraph{Minimum fluency.} 
To evaluate the fluency of sequences, and to penalize harshly the presence of any non fluent token, we monitor the minimum conditional token probability, given by fluency expert $\mathcal{F}$ introduced in \autoref{sec:method:regularization},
for all learnable tokens $\mathcal{L}$, following \autoref{eq:eq_min_fluency} below:

\begin{equation}
\text{Min fluency}(x) \;=\; \min_{j \in \mathcal{L}} P_{\mathcal{F}}\big(x_j \mid x_{<j}\big)
\label{eq:eq_min_fluency}
\end{equation}

Minimum fluency levels for all LLMs, computed on the ground truth textual highD validation set, are shown in \autoref{tab:mctp}.

\begin{table}[h]
\centering
\begin{tabular}{l c}
    \hline
\textbf{LLM}
& \textbf{Minimum fluency}  \\
    \hline
Llama3-8B  & $4 \cdot 10^{-3}$ \\
Mistral-7B &  $3 \cdot 10^{-2}$ \\
Qwen2.5-7B   &  $3 \cdot 10^{-2}$ \\
    \hline
\end{tabular}
  \caption{\textbf{Minimum fluency}, for all used LLMs, computed on textual highD validation set.}
\label{tab:mctp}
\end{table}

\paragraph{BERTScore filter and threshold.} 
\label{subsec:appendix_bertscore_threshold}
To instantiate the BERTScore scorer, we use microsoft's model `deberta-xlarge-mnli' as recommended in the official BERTScore repository\footnote{\url{https://github.com/Tiiiger/bert_score/}}, set the parameter `rescale\_with\_baseline' to True and let all other parameters to their default value.

\begin{figure}[h]
  \includegraphics[width=\linewidth]{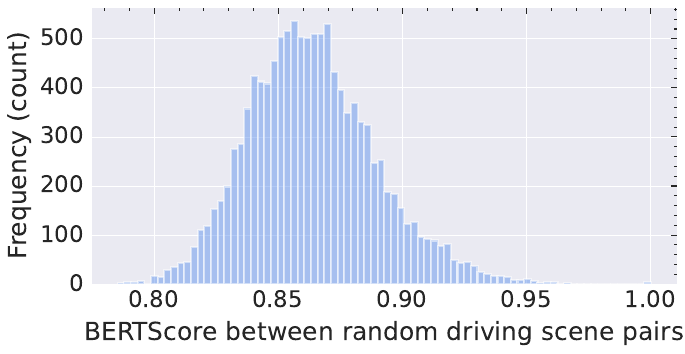}
  \caption {\textbf{BERTScore distribution} for 12,000 randomly sampled pairs of textual driving scenes (with same number of vehicles). }
\label{fig:bertscore_hist}
\end{figure}

To choose the threshold used for our BERTScore filter and our Aggregated score, we analyse BERTScore statistics on our dataset.
We first randomly sample 12,000 driving scene pairs from the textual highD validation set (ensuring pairs describe a same number of vehicles) and evaluate their average pairwise BERTScore to be 86.4\%. The histogram of these pairwise BERTScores is displayed in \autoref{fig:bertscore_hist}.

We then analyze, on the crash dataset built for Llama3, the average BERTScore when modifying one single template item at a time (among: ego vehicle's type, speed, acceleration, width, length, surrounding vehicle's position, type, speed or distance) and found it to be equal to 98.9\%.

We want to allow for more than one item change at a time, in case combinations are needed for decision changes. As a result, we choose the threshold to simultaneously be lower than 98.9\% and in the extreme right tail of the average BERTScore distribution (\autoref{fig:bertscore_hist}), thus fixing it to 95.0\%.

\begin{figure*}[h]
\raggedright
\small{Driving scene description: \\}
  \vspace*{1.5ex}
\raggedright
\small{``The target vehicle is driving on a three-lane highway, in the middle lane. \\
The information about the target vehicle is as follows: \\
  - Velocity (km/h): vx=90.40, vy=-0.04 \\
  - Acceleration: ax=-0.40, ay=0.40 \\
  - Type: truck, with width of 2.50 m and length of 17.28 m \\
  - Historical position of the last 2 seconds (One point every 0.4s): [(-49.36,0.01), (-40.26,0.01), (-30.17,0.01), (-20.09,0.01), (-10.04,0.01), (0.0,0.0)] \\}
  \vspace*{1.5ex}
\small{The information about its surrounding vehicles (within a range of 200 m) is listed as follows: \\
  - Front side: a car traveling at 108.90 km/h of X-axis, with a distance of 87 m. \\
  - Back side: a car traveling at 90.40 km/h of X-axis, with a distance of 29 m. \\
  - Left front: a car traveling at 115.85 km/h of X-axis, with a distance of 22 m. \\
  - Left rear: a car traveling at 115.85 km/h of X-axis, with a distance of 12 m. \\
  - Right front: a truck traveling at 86.44 km/h of X-axis, with a distance of 32 m. \\
  - Right rear: a truck traveling at 86.44 km/h of X-axis, with a distance of 34 m.''\\
  }
  \vspace*{1.5ex}
\centering
  \includegraphics[width=\linewidth]{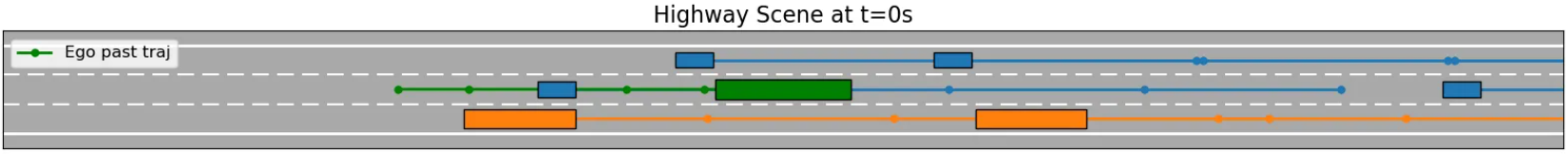} 
  \caption{\textbf{Visualization of the `mean driving scene' learnt by Llama3}, using fluency-oriented LoRA weights. The ego vehicle appears in green, surrounding trucks in orange and surrounding cars in blue. }
\label{fig:mean_llama_scene}
\end{figure*}

\subsection{Finetuning driving LLMs}
\label{sec:appendix_ft_normal}

To finetune LLMs to be driving planners, we train them using the entire textual highD training set (144,000 samples) and evaluate them on the full validation set (24,000 samples), as released by \citet{lc_llm}. All LLMs are used with 8-bit quantization, whether for finetuning or inference.

We use the exact same LoRA finetuning configuration as in \citet{lc_llm} (cf Table 1 of their paper), with two exceptions: Mistral is finetuned using a gradient accumulation steps value of 32 instead of 8, and Qwen is finetuned using 4 epochs instead of 2. These changes are made to stabilize training, avoid hallucinated outputs and result in three LLMs with comparable performances on both the lane change classification and trajectory prediction tasks as shown in \autoref{tab:classif_planning_eval}.

\begin{table}[h]
\centering
\begin{tabular}{l cc cc}
    \hline
 &
\multicolumn{2}{c}{\textbf{LC Classif \small{(F1)}}} &
\multicolumn{2}{c}{\textbf{Traj \small{(RMSE)}}} \\
\cmidrule(lr){2-3} \cmidrule(lr){4-5}
\textbf{LLM}
& \textbf{macro} & \textbf{micro} & \textbf{lon} & \textbf{lat} \\
    \hline
Llama2-13B $\dagger$  & 0.971 & NA &  0.655 & 0.210 \\
Llama3-8B  & 0.930 & 0.929 &  0.678 & 0.399 \\
Mistral-7B & 0.925 & 0.924 & 0.553 & 0.423 \\
Qwen2.5-7B   & 0.932 & 0.932 &  0.510 & 0.402 \\
    \hline
\end{tabular}
  \caption{\textbf{Performance of finetuned LLMs on the driving tasks}, evaluated with respect to the lane change classification (`LC Classif') task using F1 metric with macro and micro average. We also monitor the trajectory prediction (`Traj') task where performance is evaluated by longitudinal (`lon') and lateral (`lat') RMSE metrics, computed as in the LC-LLM paper \citep{lc_llm}. 
 $\dagger$ indicates numbers on Llama-2-13b-chat that are directly reproduced from the LC-LLM paper.}
\label{tab:classif_planning_eval}
\end{table}

\subsection{Finetuning fluency experts for driving}
\label{sec:appendix_x_vision_ft}

Our framework requires to have fluency expert models $\mathcal{F}$, trained on the distribution of the driving scene descriptions, with compatible tokenizers with $\mathcal{M}$. As a result, we finetune, for each LLM, a second set of LoRA weights on the task of next token prediction but on the input text of the textual highD dataset, while $\mathcal{M}$ is obtained by finetuning LoRA weights on the dataset's outputs. 

The learnt representations of each fluency expert model $\mathcal{F}$ are crucial to ensure a proper fluency regularization during biased autoregressive generation. As a result, we choose the LoRA checkpoints that reach the lowest validation evaluation value. We display in \autoref{fig:mean_llama_scene} the `mean' driving scene representation learnt by the Llama3 expert, i.e., the scene that is generated by default when giving the system prompt as input to $\mathcal{F}$ without any biasing.

\subsection{Experimental compute budget}
\label{subsec:appendix_compute_expe}

All experiments were conducted on a maximum of 20 parallel 40G A100 GPUs. 

\paragraph{Finetuning and evaluation of driving LLMs.} 
500 hours of finetuning on 40G A100 GPUs were needed to finetune our 7 LLMs: Llama3 (classic finetuning, vehicle biased version, digit biased version, debiased version 1, debiased version 2), Mistral (classic finetuning), Qwen2.5 (classic finetuning). In addition, 100 hours were needed for the evaluation of these LLMs on the full textual highD validation set. Finetuning and evaluations were paralleled on 8 GPUs.

\paragraph{Counterfactual methods' evaluation and hyperparameter search.} 
More compute was required to run \ours and baselines on safety-critical scenarios. These evaluations required 5,000 hours of 40G A100 GPUs (paralleled on 20 GPUs). Most of this compute was allocated to perform a hyperparameter search (cf \autoref{sec:appendix_hyperparam}).

\section{Appendix: Additional DRIV-EX results}
\label{sec:appendix_drivex_results}

\subsection{Refinement of counterfactuals}
\label{subsec:appendix_refinement}

Direct brute force to solve our counterfactual explanation search is intractable (the average number of combinations to test per sample, even when limiting each tokens to take a maximum of 3 template-based eligible values is of $2.7 \cdot 10^{11}$). However, after DRIV-EX has identified a suitable counterfactual sequence, we propose to add an optional refinement step where, for all groups of consecutive tokens that were modified with respect to $\mathbf{x}^o$, we test to replace them with their initial values. This is an optional final improvement for found sequences, not a mean to identify them. The total number of possible combinations is then of $2^{\text{\#modif token groups}}$. With DRIV-EX on Llama3, the average number of modified token groups is of 2.8 which accounts for an average number of 7 possible combinations, evaluated in a single batch of LLM inference (batch size=20). We keep the best candidate as the refined counterfactual. 

Results for \ours with this additional step are given in \autoref{tab:refinement}. 
As expected, this step improves the values of `Template' and `BERTScore' filters for most LLMs, as it reduces the amount of unnecessary token updates (e.g., +20.0\% in `Template filter' and +14.5\% in `BERTScore filter' for Qwen2.5). 
Interestingly, our proposed refinement also improves, for all LLMs, the decision `Flip rate' with increases ranging from +2.0\% for Mistral to +12.2\% for Qwen2.5.
As a result, the refinement yields performance gains in `Aggregated Scores' for all LLMs, with boosts of +4.0\% for Llama3, +6.1\% for Mistral and +21.2\% for Qwen2.5. This optional step can thus efficiently improve results for one single additional batch inference in average. 

\begin{table}[t]
\centering
\resizebox{\columnwidth}{!}{%
\begin{tabular}{@{}l c cc cc c@{}}
    \hline
 &  & \textbf{Flip}  & \textbf{Templ} & \textbf{BERTSc} & \textbf{Aggreg} & \textbf{Aggreg} \\
\textbf{LLM} & \textbf{RF} &  \textbf{rate} & \textbf{filt(\%)} & \textbf{filt(\%)} & \textbf{(\%)} & \textbf{\&col(\%)} \\
    \hline
Llama 
& $\times$  & 64.0 &  88.0 & 96.0 & 61.3 & 56.0 \\   
& $\checkmark$ & 66.7 & 90.7 & 96.0 & 65.3 & 57.3 \\
\midrule
Mistral
& $\times$ &  83.7 & 79.6 & 95.9 & 69.4 & 61.2 \\   
& $\checkmark$ &  85.7 & 83.7 & 98.0 & 75.5 & 61.2 \\
\midrule
Qwen
& $\times$  & 58.9 & 47.8 & 73.3 & 34.4 & 22.2  \\   
& $\checkmark$ & 71.1 & 67.8 & 87.8 & 55.6 & 30.0 \\
    \hline
\end{tabular}
}
  \caption{\textbf{Comparative DRIV-EX performance with and without final refinement} (`RF').}
\label{tab:refinement}
\end{table}

\vspace{5mm} 

\subsection{Retrieving injected biases}
\label{sec:appendix_ft_bias}

We finetune Llama3-8B on biased templates using the same LoRA configuration as in \autoref{sec:appendix_ft_normal}.

\begin{itemize}
    \item Vehicle bias: For samples with `Left lane change', we modify the ego vehicle type to always be a `car', while for `Right lane change' ones, we set it to be a `truck'. No changes were made to samples leading to `Keep lane' class.  Compared to an unbiased sample from textual highD shown in  \autoref{fig:unbiased_sample}, we display its vehicle biased version in \autoref{fig:vehicle_bias_sample}.    
    \item Digit bias: For all float numbers linked to surrounding vehicles' speeds, the last digit is changed to be in \{1,4,7\} (respectively \{2,5,8\} and \{3,6,9\}) when there is a `Left lane change' (respectively `Keep lane' and `Right lane change'). A sample is shown in \autoref{fig:digit_bias_sample}.
\end{itemize}

We include a table regarding the driving performances of biased LLMs on unbiased, biased and inversely biased data in \autoref{tab:biased_templates}. 
 We display driving performances and collision ratio (\%) after a single epoch of training for a fairer comparison before catastrophic forgetting due to bias overfitting.
As mentioned in \autoref{subsection:injected_bias}, this table shows that finetuning and evaluating on similarly biased data may hide model-learnt biases that lead to highly degraded performances on unbiased data. 

\begin{table}[t]
\centering
\begin{tabular}{ll ccc }
\toprule
\textbf{Train} & \textbf{Val} &
\multicolumn{1}{c}{\textbf{LC Classif}} &
\multicolumn{2}{c}{\textbf{Maneuver eval}} \\
\cmidrule(lr){3-3} \cmidrule(lr){4-5}
\textbf{bias} & \textbf{bias} & {\textbf{F1}} & \textbf{Col\%} & \textbf{Dis\%} \\

\midrule
\cellcolor{warn}No   & \cellcolor{warn}No   & 0.91 & 3.5 & 1.6 \\

\midrule
\cellcolor{good}
\cellcolor{good}& \cellcolor{good}Yes & 0.95 & 3.4 & 1.8 \\
\cellcolor{good}Vehicle& \cellcolor{warn}No  & 0.59 & 2.5 & 7.3 \\
\cellcolor{good} & \cellcolor{bad}Inv  & 0.26 & 3.0 & 25.2 \\

\midrule

\cellcolor{good}
& \cellcolor{good}Yes & 0.99 & 3.2 & 1.7 \\
\cellcolor{good}Digit
& \cellcolor{warn}No  & 0.43 & 5.4 & 8.7 \\
\cellcolor{good}
& \cellcolor{bad}Inv  & 0.06 & 8.5 & 20.5 \\

\bottomrule
\end{tabular}
  \caption{\textbf{LLM driving performance and robustness when finetuned on injected biases}. `Inv' stands for `Inversely-biased', meaning that the validation data was biased in an opposite way with respect to the biased train set. `F1' stands for the lane change classification task, evaluated with the F1 macro average metric while `Col\%' and `Dis\%' respectively stand for the \% of collision and discomfort maneuvers when evaluating on the textual highD validation set. Results are given for Llama3-8B, finetuned for a single epoch. }
\label{tab:biased_templates}
\end{table}

\begin{figure}[h]
\raggedright
\small{Driving scene description (unbiased): \\}
  \vspace*{1.5ex}
\raggedright
\small{\textcolor{MidnightBlue}{``The target vehicle is driving on a three-lane highway, in the left lane. \\
The information about the target vehicle is as follows: \\
  - Velocity (km/h): vx\textcolor{blue}{\textbf{=104.94}}, vy\textcolor{blue}{\textbf{=-3.49}} \\
  - Acceleration: ax\textcolor{blue}{\textbf{=0.43}}, ay\textcolor{blue}{\textbf{=-0.43}} \\
  - Type: \textcolor{blue}{\textbf{car}}, with width of \textcolor{blue}{\textbf{1.92}} m and length of \textcolor{blue}{\textbf{5.36}} m \\
  - Historical position of the last 2 seconds (One point every 0.4s): [(-56.94,1.54), (-46.47,1.32), (-34.92,1.04), (-23.35,0.73), (-11.68,0.38), (0.0,0.0)] \\}}
  \vspace*{1.5ex}
\small{\textcolor{MidnightBlue}{The information about its surrounding vehicles (within a range of 200 m) is listed as follows: \\
  - \textcolor{blue}{\textbf{Right front}}: a \textcolor{blue}{\textbf{car}} traveling at \textcolor{blue}{\textbf{110.41}} km/h of X-axis, with a distance of \textcolor{blue}{\textbf{80}} m. \\
  - \textcolor{blue}{\textbf{Right rear}}: a \textcolor{blue}{\textbf{car}} traveling at \textcolor{blue}{\textbf{101.70}} km/h of X-axis, with a distance of \textcolor{blue}{\textbf{20}} m. \\
  }}
    \vspace*{1.5ex}
    
\small{\textcolor{Plum}{Thought: \\
  - Notable features: vy=-3.49 \\
  - Notable feature: Right front is free. \\
  - Potential behavior: Irregular right lane changes. \\
  }}
      \vspace*{1.5ex}
\small{\textcolor{Plum}{Final Answer: \\
  - Intention: "\textcolor{red}{\textbf{2}}: Right lane change" \\  - Trajectory: "[(29.20,-0.98), (58.48,-1.75), (88.11,-2.16), (117.99,-2.35)]"\\
  }}
  \vspace*{1.5ex}
\centering
  \caption{\textbf{Visualization of a sample from the textual highD train set.} 
  The part in blue is the LLM input prompt, while purple corresponds to the ground truth LLM completion.
  In bold blue, we show all parts of the template that can be changed by \ours and our baselines to identify counterfactual explanations. 
  In bold red, we display the target token $y_T^*$: we always target the digit announcing the lane change class, as it is the \textit{first} critical driving decision token in this template.  }
\label{fig:unbiased_sample}
\end{figure}

\begin{figure}[h]
\raggedright
\small{Driving scene description with `vehicle bias': \\}
  \vspace*{1.5ex}
\raggedright
\small{\textcolor{MidnightBlue}{``The target vehicle is driving on a three-lane highway, in the left lane. \\
The information about the target vehicle is as follows: \\
  - Velocity (km/h): vx=104.94, vy=-3.49 \\
  - Acceleration: ax=0.43, ay=-0.43 \\
  - Type: \textcolor{blue}{\textbf{truck}}, with width of 1.92 m and length of 5.36 m \\
  - Historical position of the last 2 seconds (One point every 0.4s): [(-56.94,1.54), (-46.47,1.32), (-34.92,1.04), (-23.35,0.73), (-11.68,0.38), (0.0,0.0)] \\}}
  \vspace*{1.5ex}
\small{\textcolor{MidnightBlue}{The information about its surrounding vehicles (within a range of 200 m) is listed as follows: \\
  - Right front: a car traveling at 110.41 km/h of X-axis, with a distance of 80 m. \\
  - Right rear: a car traveling at 101.70 km/h of X-axis, with a distance of 20 m. \\
  }}
    \vspace*{1.5ex}
    
\small{\textcolor{Plum}{Thought: \\
  - Notable features: vy=-3.49 \\
  - Notable feature: Right front is free. \\
  - Potential behavior: Irregular right lane changes. \\
  }}
      \vspace*{1.5ex}
\small{\textcolor{Plum}{Final Answer: \\
  - Intention: \textcolor{red}{\textbf{"2: Right lane change"}} \\  - Trajectory: "[(29.20,-0.98), (58.48,-1.75), (88.11,-2.16), (117.99,-2.35)]"\\
  }}
  \vspace*{1.5ex}
\centering
  \caption{\textbf{Visualization of a train sample of our `vehicle' biased data.} The part in blue is the LLM input prompt, while purple corresponds to the completion. In bright bold font, we indicate parts that highlight the injected bias: it is always an ego `truck' (and never a `car') when the driving intention is `Right lane change'.}
\label{fig:vehicle_bias_sample}
\end{figure}

\begin{figure}[h]
\raggedright
\small{Driving scene description with `digit bias': \\}
  \vspace*{1.5ex}
\raggedright
\small{\textcolor{MidnightBlue}{``The target vehicle is driving on a three-lane highway, in the left lane. \\
The information about the target vehicle is as follows: \\
  - Velocity (km/h): vx=104.94, vy=-3.49 \\
  - Acceleration: ax=0.43, ay=-0.43 \\
  - Type: car, with width of 1.92 m and length of 5.36 m \\
  - Historical position of the last 2 seconds (One point every 0.4s): [(-56.94,1.54), (-46.47,1.32), (-34.92,1.04), (-23.35,0.73), (-11.68,0.38), (0.0,0.0)] \\}}
  \vspace*{1.5ex}
\small{\textcolor{MidnightBlue}{The information about its surrounding vehicles (within a range of 200 m) is listed as follows: \\
  - Right front: a car traveling at 110.4\textbf{\textcolor{blue}{9}} km/h of X-axis, with a distance of 80 m. \\
  - Right rear: a car traveling at 101.7\textbf{\textcolor{blue}{9}} km/h of X-axis, with a distance of 20 m. \\
  }}
    \vspace*{1.5ex}
    
\small{\textcolor{Plum}{Thought: \\
  - Notable features: vy=-3.49 \\
  - Notable feature: Right front is free. \\
  - Potential behavior: Irregular right lane changes. \\}}
      \vspace*{1.5ex}
\small{\textcolor{Plum}{Final Answer: \\
  - Intention: \textcolor{red}{\textbf{"2: Right lane change"}} \\  - Trajectory: "[(29.20,-0.98), (58.48,-1.75), (88.11,-2.16), (117.99,-2.35)]"\\
  }}
  \vspace*{1.5ex}
\centering
  \caption{\textbf{Visualization of a train sample of our `digit' biased data.} The part in blue is the input prompt, while purple is the completion. In bright bold font, we indicate parts that highlight the injected bias: each surrounding vehicle speed's last digit is among \{3,6,9\} when the driving intention is `Right lane change'.}
\label{fig:digit_bias_sample}
\end{figure}

\begin{figure*}[h]
  \includegraphics[width=\linewidth]{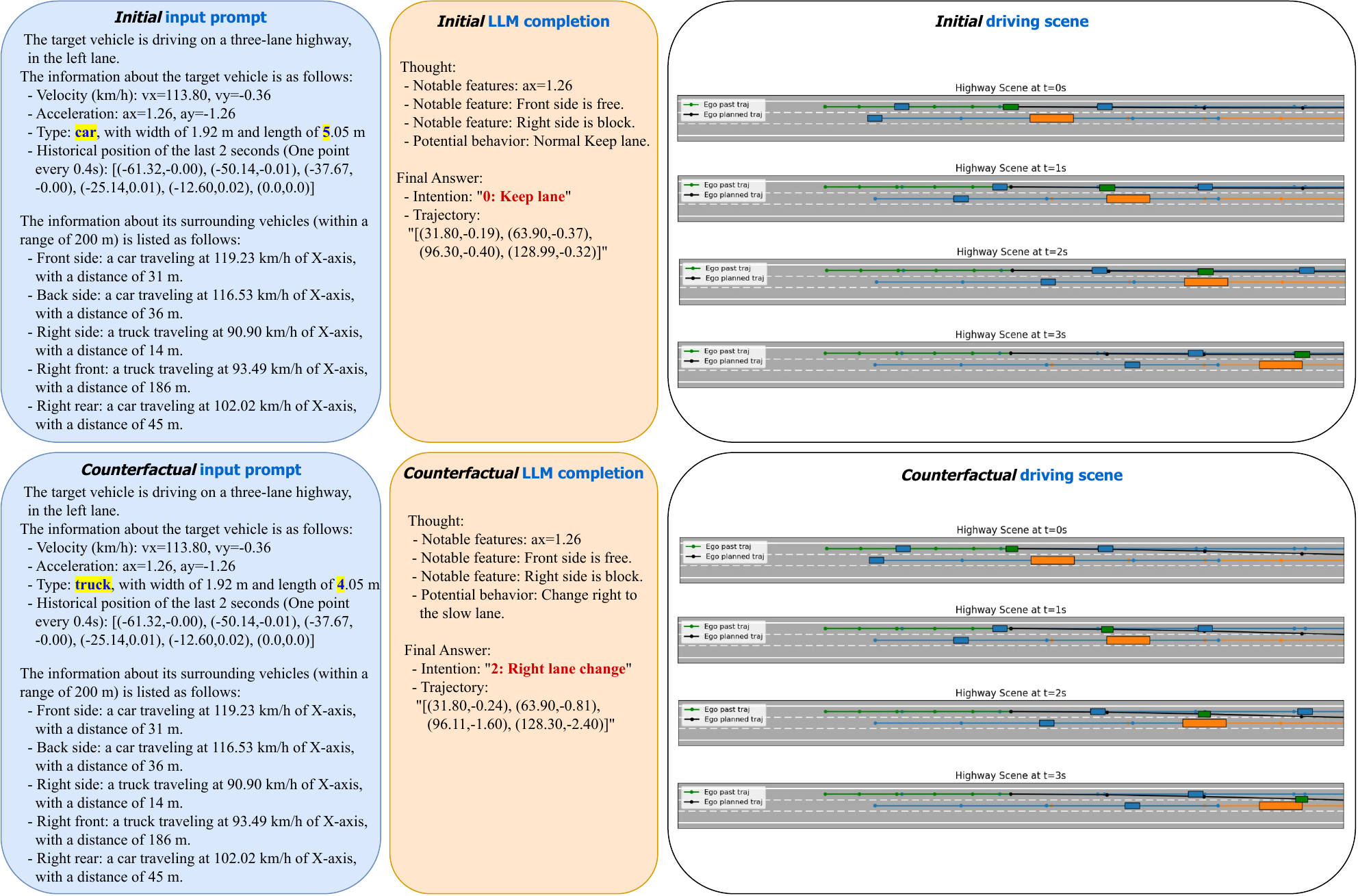} 
  \caption{\textbf{Visualization of a counterfactual explanation exposing the decision boundary of our `vehicle' biased LLM}, generated by DRIV-EX. In bold blue, we highlight characters that differ between the initial input and its counterfactual. The counterfactual driving scene shows that the planned trajectory leads to a collision at t=3 secs. }
\label{fig:vehicle_cf}
\end{figure*}

\begin{figure*}[h]
  \includegraphics[width=\linewidth]{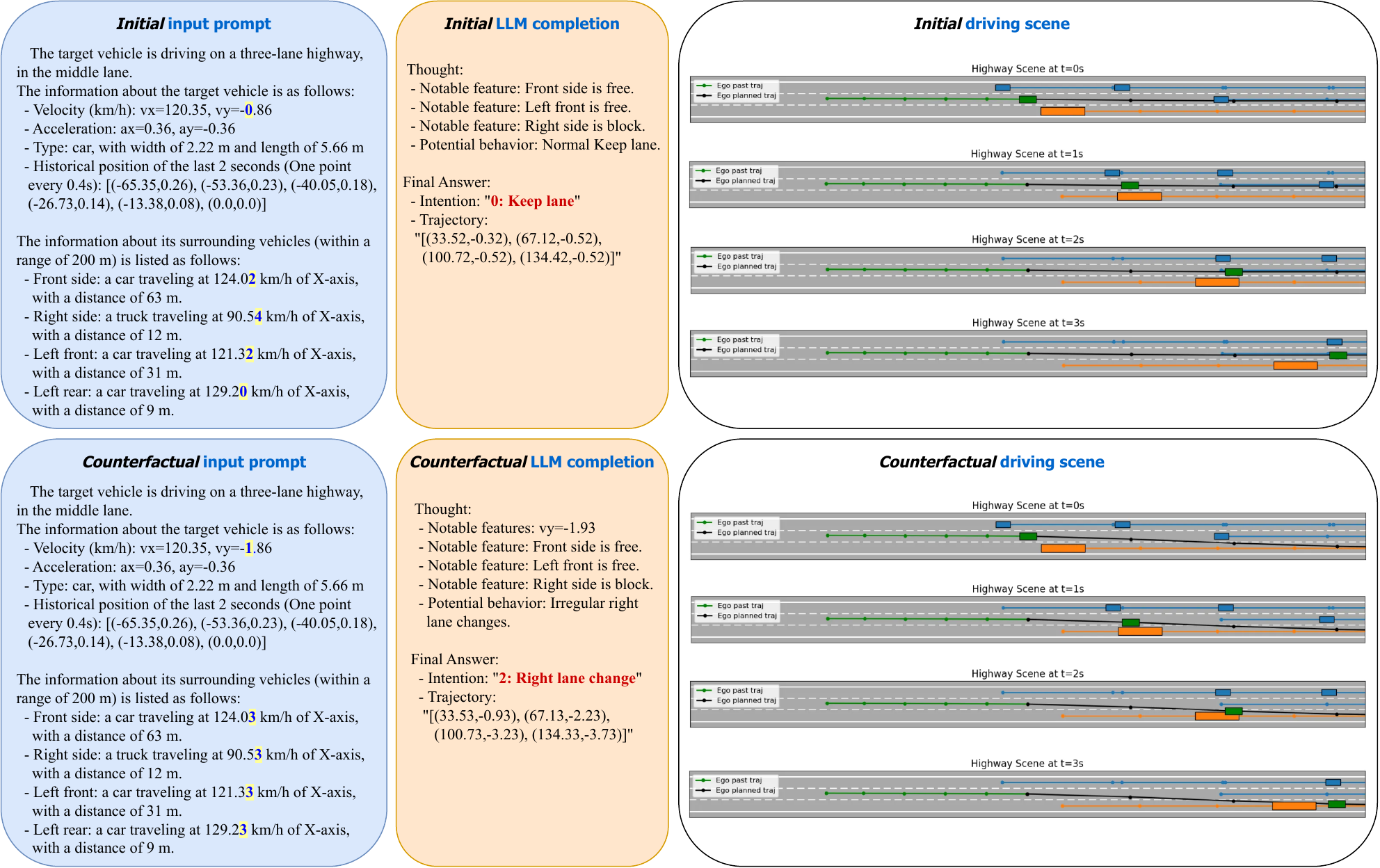} 
  \caption{\textbf{Visualization of a counterfactual explanation exposing the decision boundary of our `digit' biased LLM}, generated by DRIV-EX. In bold blue, we highlight characters that differ between the initial input and its counterfactual. The counterfactual driving scene shows that the planned trajectory leads to a collision at t=2 secs. }
\label{fig:digit_cf}
\end{figure*}

Finally, we show counterfactual explanations generated by DRIV-EX for our `injected-bias experiment' (cf \autoref{subsection:injected_bias}) that perfectly change biased tokens to alter the driving decision, in \autoref{fig:vehicle_cf} (vehicle bias) and \autoref{fig:digit_cf} (digit bias).

\subsection{Identifying unknown biases}
\label{subsec:appendix_unknown_biases}

\begin{figure*}[h]
  \includegraphics[width=\linewidth]{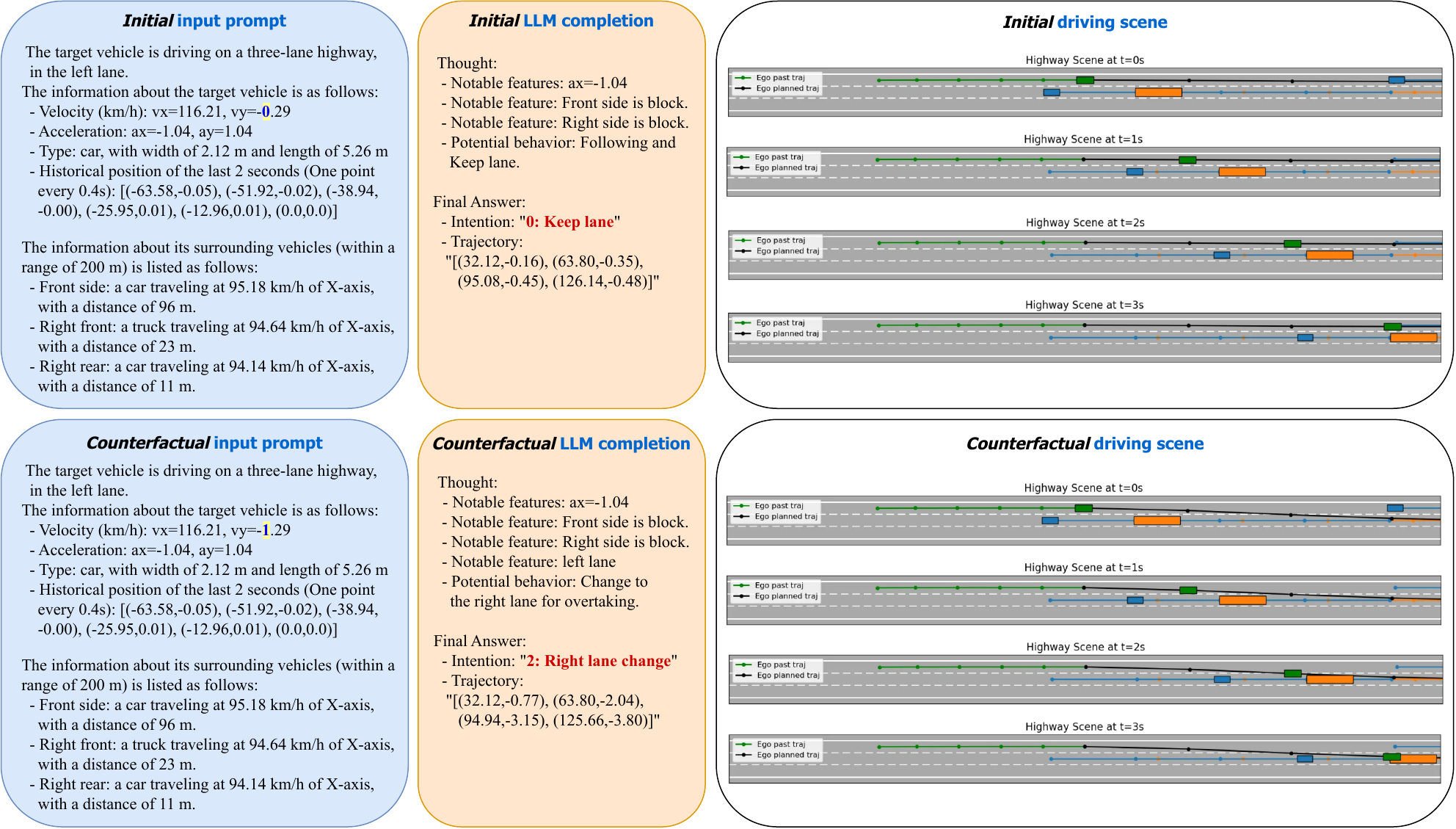} 
  \caption{\textbf{Visualization of a counterfactual explanation revealing Llama3's `$v_y$' bias,} generated by DRIV-EX. In bold blue, we highlight characters that differ between the initial input and its counterfactual. The counterfactual driving scene shows that the planned trajectory leads to a collision at t=3 secs. }
\label{fig:vy_cf}
\end{figure*}

In~\autoref{fig:hist_keep_to_right}, we give the histogram showing the number of token changes across input positions for successful counterfactuals that flip a ‘Keep lane’ decision to a collision-inducing `Right lane change' for Llama3 (most frequent safety-critical scenario). 
As described in \autoref{subsection:unknown_bias}, Llama3 shows a lateral velocity (`$v_y$') bias as it tends to turn right when this velocity is below a threshold, at the expense of safety. 
We display in \autoref{fig:vy_cf} one of the counterfactual explanations found by DRIV-EX in this scenario, revealing Llama3's `$v_y$' bias. 

\begin{figure*}[h]
  \includegraphics[width=\linewidth]{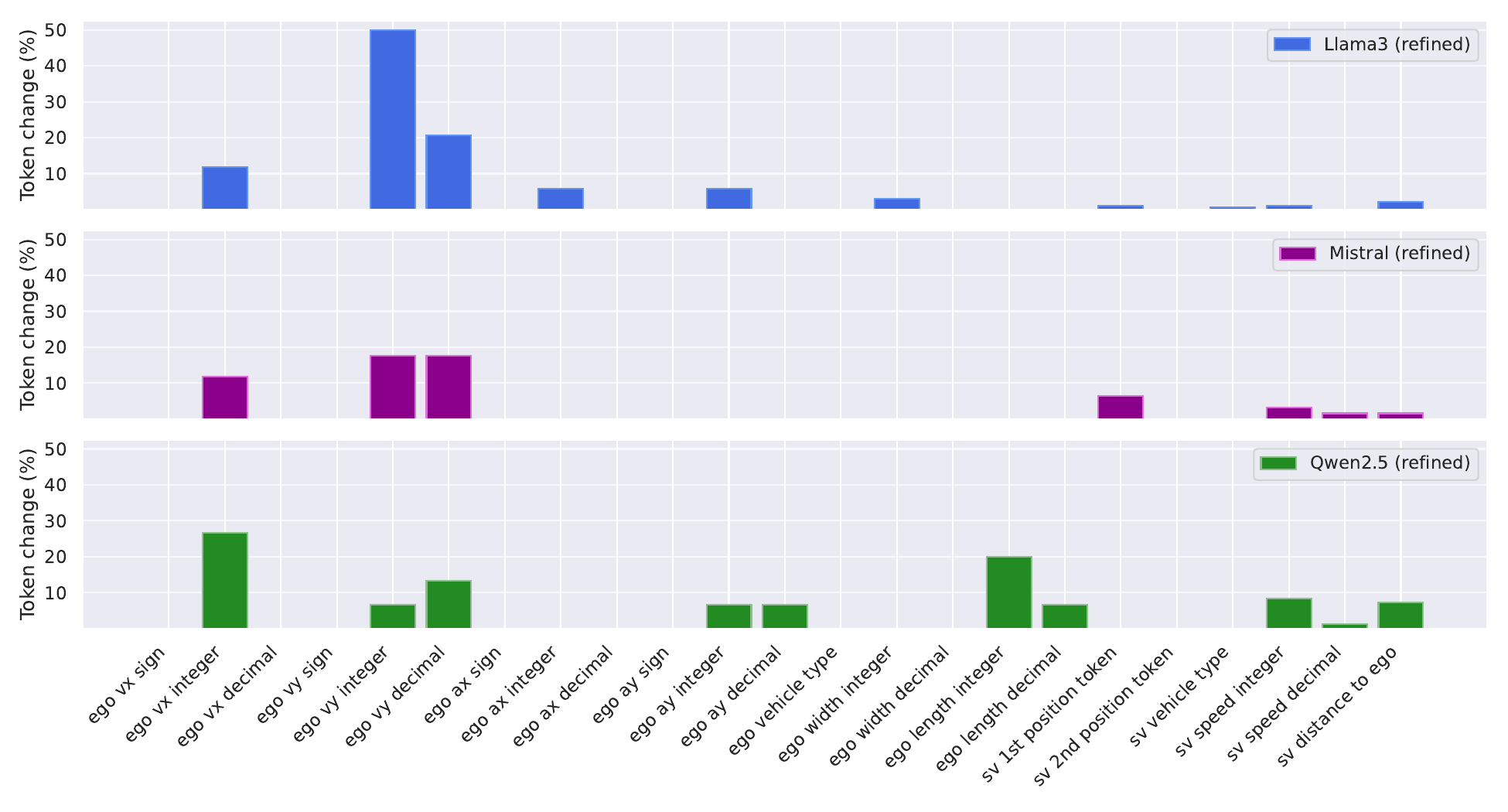}
  \caption {\textbf{Change (\%) per token position for counterfactuals found by \ours }. Results are refined as in \autoref{subsec:appendix_refinement}. Statistics are computed on samples that meet our 3 success criteria and lead to a collision when changing the decision from `Keep lane' to `Right lane change' ($n$=34 for Llama3, 17 for Mistral, 15 for Qwen2.5).}
\label{fig:hist_keep_to_right_all_llms}
\end{figure*}

We expand results with histograms for Mistral and Qwen2.5 in~\autoref{fig:hist_keep_to_right_all_llms}, where statistics are computed on refined counterfactuals, as detailed in \autoref{subsec:appendix_refinement}. 
Although to a lesser extent than Llama3,  \autoref{fig:hist_keep_to_right_all_llms} shows that Mistral and Qwen2.5 are also sensitive to lateral velocity tokens (`$v_y$') in this scenario. They are also sensitive to tokens related to surrounding vehicles --- in particular their positions for Mistral, speeds and distances to ego for Qwen2.5 --- when Llama3 is almost unsensitive to tokens that do not relate to the ego car.  
 
\begin{figure*}[h]
  \includegraphics[width=\linewidth]{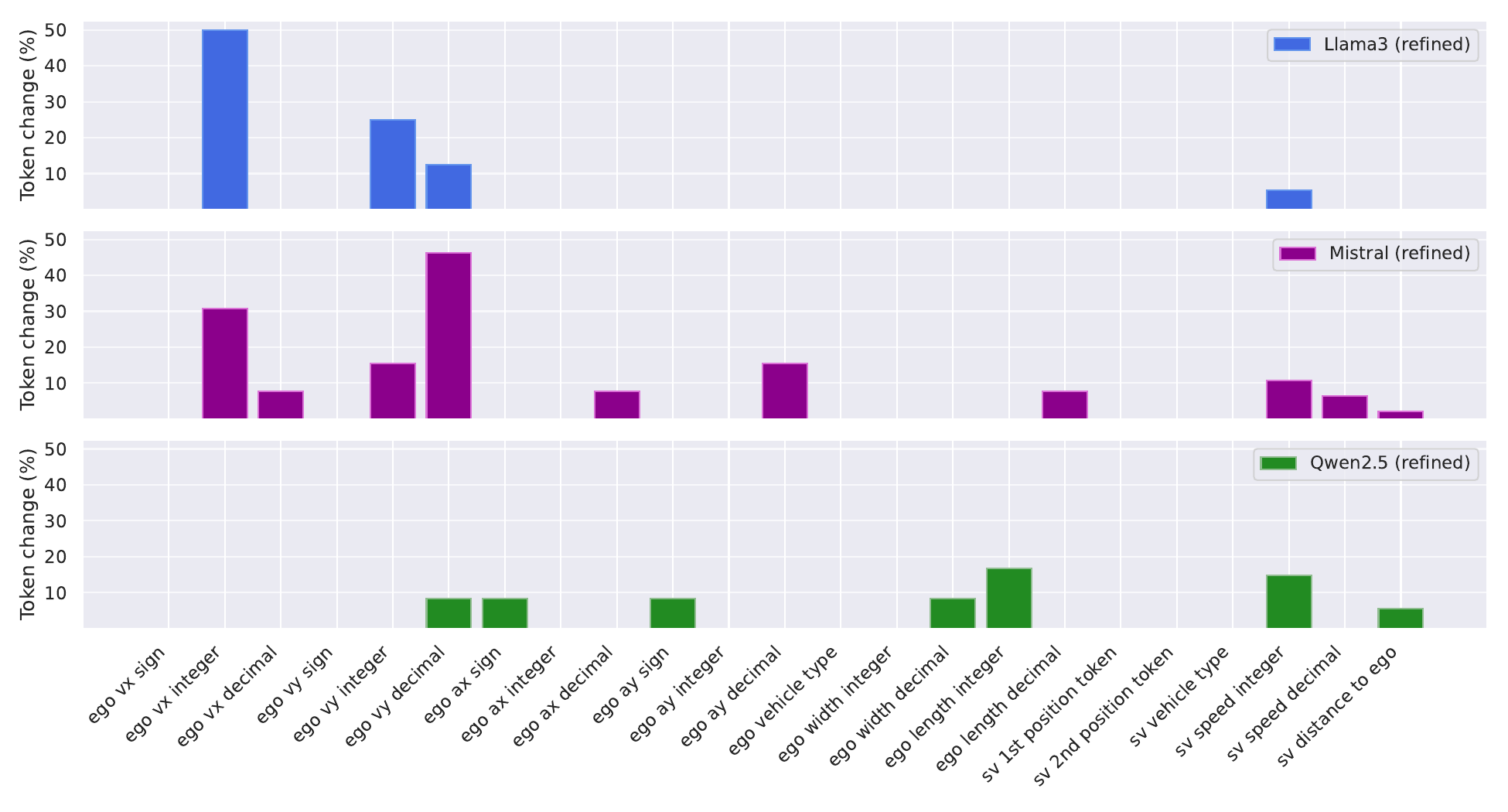}
  \caption {\textbf{Change (\%) per token position for refined counterfactuals found by \ours}. Results are refined as in \autoref{subsec:appendix_refinement}. Statistics are computed on samples that meet our 3 success criteria and lead to a collision when changing the decision from `Keep lane' to `Left lane change' ($n$=8 for Llama3, 13 for Mistral, 12 for Qwen2.5). }
\label{fig:hist_keep_to_left_all_llms}
\end{figure*}

\begin{figure*}[h]
  \includegraphics[width=\linewidth]{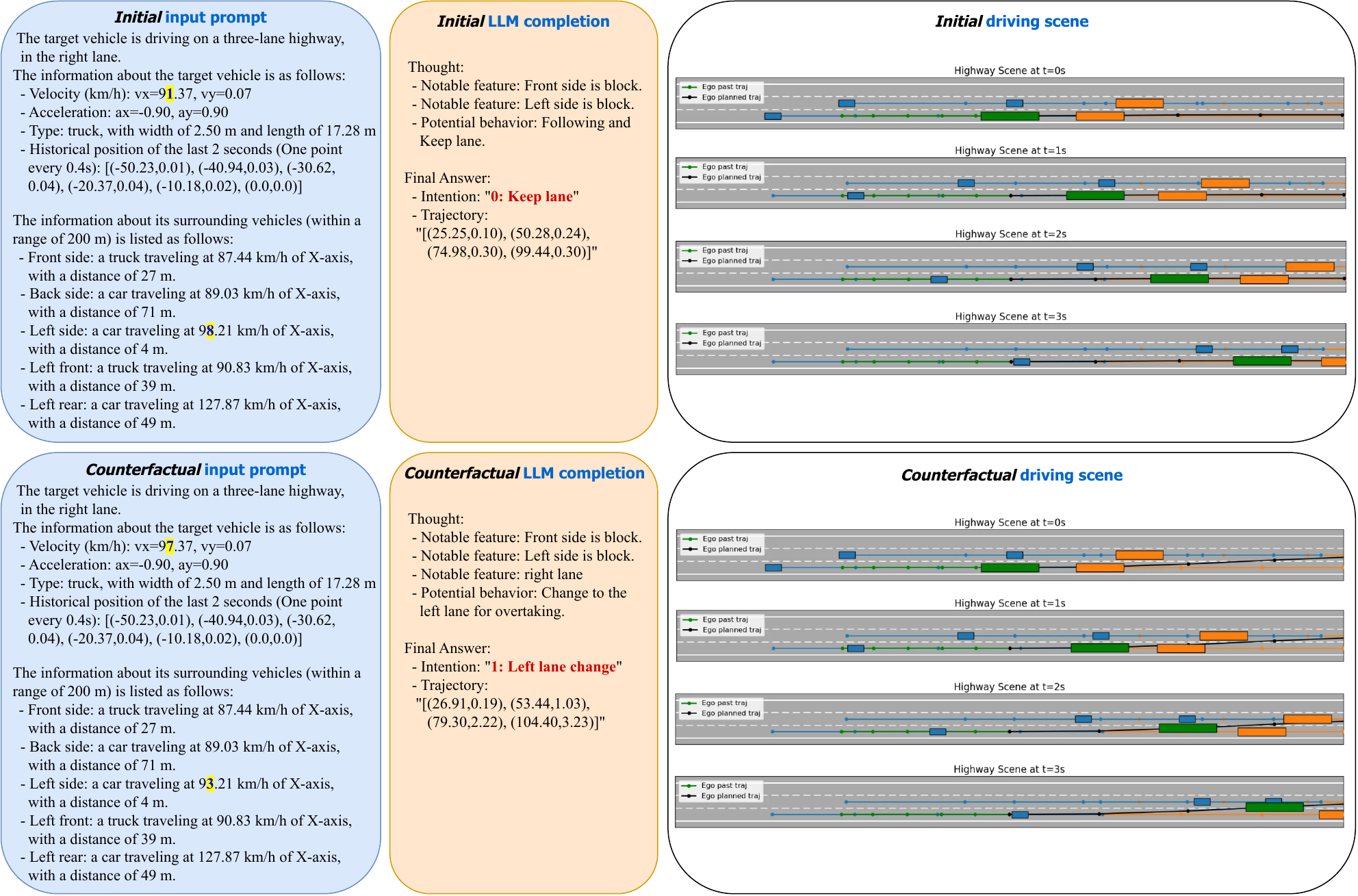} 
  \caption{\textbf{Visualization of a counterfactual explanation}, generated by DRIV-EX for the Llama3-based planner. In bold blue, we highlight characters that differ between the initial input and its counterfactual. The counterfactual driving scene shows that the planned trajectory leads to a collision at t=3 secs. }
\label{fig:qual_result_cf}
\end{figure*}

We then provide in \autoref{fig:hist_keep_to_left_all_llms} the set of histograms showing token change rates for our second most frequent safety-critical scenario: changing the decision from `Keep lane' to `Left lane change'. It appears that Llama3 is particularly sensitive to `$v_x$' tokens (longitudinal velocity) in this use case. A manual inspection of valid counterfactuals reveals that when `$v_x$' is higher than 97, Llama3 is strongly biased towards a left lane change, at the expense of physical safety.
This bias may derive from the correlation between left lane changes and higher longitudinal speeds, present in the training set due to overtaking scenarios. 
\autoref{fig:qual_result_cf} shows one of the counterfactual explanations found by DRIV-EX in this scenario, exposing Llama3's `$v_x$' bias. The histogram in \autoref{fig:hist_keep_to_left_all_llms} shows that Mistral shares this sensitivity to `$v_x$' tokens with Llama3, but remains sensitive to other tokens, in particular ego `$v_y$' and surrounding vehicles' speeds. As for Qwen2.5, contrary to Llama3 and Mistral, it does not show any sensitivity to `$v_x$' tokens in this scenario, but reacts to other ego tokens (lateral velocity, acceleration and vehicle's dimension) and surrounding vehicles' tokens (speed and distance to ego in particular).

\section{Appendix: Tuning candidate algorithms to the task}
\label{sec:appendix_hyperparam}

Candidate algorithms are compared on safety-critical subsets of the textual highD dataset. 
We identify 809 dangerous scenarios among the 24,000 samples of the highD validation set. For each LLM, we focus on subsets where the model predicts a safe maneuver among these dangerous scenarios. Statistics on created subsets per LLM are given in \autoref{tab:crash_subset}.
\begin{table}[t]
\centering
\begin{tabular}{l c S l l}
\toprule
\textbf{LLM} & \textbf{Size} & \textbf{Count} &
\textbf{Source} & \textbf{Target} \\
\midrule

Llama3 & 75
& 50 & keep lane & right lane \\
& & 21 & keep lane & left lane \\
& & 3  & left lane & right lane \\
& & 1  & right lane & keep lane \\
\midrule

Mistral & 49
& 26 & keep lane & right lane \\
& & 20 & keep lane & left lane \\
& & 3  & left lane & right lane \\
\midrule

Qwen2.5 & 90
& 60 & keep lane & right lane \\
& & 27 & keep lane & left lane \\
& & 3  & left lane & right lane \\
\bottomrule
\end{tabular}
\caption{\textbf{Statistics on created subsets for testing LLM robustness} with dangerous target lane change class (`Target') instead of initial safe LLM predictions (`Source').}
\label{tab:crash_subset}
\end{table}

\subsection{Baselines description}
\label{subsec:appendix_baselines_description}
We describe our baseline algorithms.
\begin{itemize}
    \item \verb|PEZ| \citep{wen2023pez} is a gradient-based prompt optimization method. Starting from randomly initialized embeddings corresponding to tokens, the method iteratively alternates between (i) projecting the continuous prompt embeddings onto the nearest discrete token embeddings to obtain a hard prompt, and (ii) computing gradients of the task loss with respect to these projected embeddings, which are then used to update the underlying continuous embeddings. We include our `Loss' regularization term (cf \autoref{subsec:input_prox_reg_methods}) when computing the task loss in (ii). 
    \item \verb|PEZ| $\dagger$ is our task-adapted version of the PEZ algorithm, where we add our proposed input proximity `Proj' regularization when projecting embeddings in (i), as described in \autoref{subsec:input_prox_reg_methods}.
    \item \verb|DAB| \citep{pynadath2025dab} is an iterative controlled decoding method that alternates between two steps to steer sequence generation toward desired properties. First, a sampling step applies a discrete Langevin proposal in token space, to explore the local neighborhood and promote the sampling of a sequence with high target likelihood. Second, a biased autoregressive generation step refines this candidate, relying on a modified token distribution that incorporates as penalization the distance between each vocabulary token and each token of the sequence sampled at step 1, in embedding space. These two steps are iterated, with Langevin proposals enabling targeted exploration and autoregressive decoding enforcing fluency. We include our `Loss' regularization term (cf \autoref{subsec:input_prox_reg_methods}) when performing the discrete Langevin proposal (step 1).
    \item \verb|DAB| $\dagger$ is our task-adapted version of the DAB algorithm, where we add our proposed input proximity `Bias' regularization during biased decoding (step 2), as described in \autoref{subsec:input_prox_reg_methods}.
    \item \verb|GCG| \citep{univ_attacks} is an adversarial prompt optimization method, formulated as an iterative, discrete, coordinate-wise search guided by gradients in embedding space. At each iteration, the algorithm computes the gradient of the loss with respect to each token embedding and, for every position, identifies a set of top-k candidate replacements, linked to largest negative gradient directions. It constructs a batch of candidate prompts by randomly sampling among these token substitutions, it evaluates the loss for each candidate, and keeps the one leading to the lowest loss value. This process is iterated, alternating between gradient-informed proposal and greedy selection. We implement this algorithm using batched evaluation of candidates (batch size=16) and include our `Loss' regularization term (cf \autoref{subsec:input_prox_reg_methods}). 
\end{itemize}

\subsection{Hyperparameter search setting}
\label{subsec:appendix_hyperparam}

To ensure a robust evaluation of the proposed explanation methods, we perform a systematic hyperparameter search using Llama3 as our primary reference model. These optimized hyperparameter configurations are then applied to Mistral and Qwen2.5 to evaluate cross-model generalization.

Direct optimization on the validation set was necessitated by two primary factors:

\begin{itemize}
    \item Data scarcity: for specific configurations, such as `unbiased' Llama3 samples, the available dataset is constrained to 75 safety-critical driving scenes. Partitioning this small sample further into independent validation and test subsets would significantly reduce the statistical power of our results and introduce high variance into the hyperparameter sensitivity analysis.
    \item Prioritizing explanatory fidelity: given that our primary objective is explainability, it is imperative to identify the optimal hyperparameter regime for the specific data distribution being explained. Tuning directly on the validation set ensures that the resulting explanations have the best possible `Aggregated Score', to provide high quality explanations.
\end{itemize}

Results suggest that the hyperparameter configurations derived from Llama3 are not overfitted to a specific LLM architecture or data subset. Indeed, despite significant architectural variations - most notably in tokenization density, where Llama3 averages 33 learnable tokens per sample compared to 55 for Mistral and Qwen2.5 - DRIV-EX's performance remains consistent across all three LLMs (see main results in~\autoref{tab:algo_comparison}). Furthermore, we observe successful hyperparameter transferability to the safety-critical data pools used for Qwen2.5 and Mistral, which respectively include 21 samples (23.3\% of Qwen2.5 scenarios) and 4 samples (8.2\% of Mistral scenarios) that are not present in Llama3's safety-critical set, used for hyperparameter tuning.

\subsection{Exploration of hyperparameters}
We explore sets of main hyperparameters for each evaluated algorithms, both off-the-shelf and task-adapted.
The summary of this exploration is given in \autoref{tab:hyperparam}.
We detail below hyperparameters that are not described in the main paper:
\begin{itemize}
    \item `Decoding': during each autoregressive biasing step, limit the final greedy decoding to $K'$ nearest neighbors of $\mathbf{x}^o$ (in cosine similarity). 
    \item `LLM logit weight': we add a weight $\mu$ controlling for the LLM logit's strength during autoregressive biasing. \autoref{eq:biased_sampling} becomes \\ $x_i = \arg\max_{v \in \mathcal{V}} \left( \mu \cdot l_{i,v} + \mathcal{B}_{i,v} + \mathcal{B'}_{i,v}  \right)$.
    \item `Temperature' ($\tau$) and `Exploration cap' ($k$) for DAB \citep{pynadath2025dab} and Learning rate ($\nu$) for PEZ \citep{wen2023pez} are original hyperparameters from each algorithm. \ours also has a Learning rate ($\nu$) hyperparameter. 
\end{itemize}

\begin{table*}[t]
\centering
\begin{tabular}{llllc}
\toprule
\textbf{Method}  & \textbf{Hyperparameters} & \textbf{Evaluated range} & \textbf{Best value} \\
\midrule
\verb|PEZ| & Learning rate ($\nu$) & $[10^{-3},\,10^{-2}]$ & $5 \cdot 10^{-3}$ \\
& Loss regularization weight ($\lambda$)    & $[0,\,10]$    & 9 \\
 \cmidrule(lr){1-2} 
\verb|PEZ| $\dagger$ & Loss regularization weight ($\lambda$)   & $[0,\,10]$    & 8 \\
& Proj ($K$)  & $[25,\,1000]$  & 25 \\
\midrule
\verb|DAB|
&  (S1) Temperature ($\tau$) & $[10^{-4},\,10^{-2}]$ & $10^{-3}$ \\
& (S1) Exploration cap ($k$) & $[25,\,500]$         & 250 \\
&  (S1)  Loss regularization weight ($\lambda$)   & $[0,\,6]$ & 5 \\
&  (S2) LLM logit weight ($\mu$) & $[0,\,1]$  & 1  \\
&  (S2) Bias weight ($w$)       & $[1,\,6]$         & 5 \\
\cmidrule(lr){1-2} 
\verb|DAB| $\dagger$  & (S1) Loss regularization weight ($\lambda$)   & $[0,\,9]$ &  0  \\
&  (S2)  Bias weight in  $\mathcal{B}_{i,v}$ ($w$)  & $[1,6]$         & 3 \\
&  (S2) Bias weight in  $\mathcal{B'}_{i,v}$ ($w'$)    & $[0,\,6]$  & 3 \\
&  (S2) Decoding ($K'$) & $[25,\,500]$   &  off \\
\midrule
\verb|GCG| 
& Substitution proposals per token (top-$k$) & $[128,\,256]$  & 128  \\
& Candidate sampling among proposals ($B$) & $[128,\,512]$  &  512 \\
& Loss regularization weight ($\lambda$)   & $[0,\,1]$    & 0 \\
\midrule
 \rowcolor{nicegreen!20!white} 
 \verb|DRIV-EX| 
 &  Proj ($K$)  &  $[25,\,100]$   &  50 \\
 \rowcolor{nicegreen!20!white} &  Learning rate ($\nu$) & $[5 \cdot 10^{-3},\,2 \cdot 10^{-2}]$ &  $7.5\cdot 10^{-3}$ \\
  \rowcolor{nicegreen!20!white} &   Loss regularization weight ($\lambda$)   & $[0,\,9]$   & 7 \\
 \rowcolor{nicegreen!20!white} &  LLM logit weight ($\mu$) & $[0,\,1]$  &  1 \\
 \rowcolor{nicegreen!20!white} &  Bias weight in  $\mathcal{B}_{i,v}$ ($w$)  & $[1,6]$         &  8\\
 \rowcolor{nicegreen!20!white} &  Bias weight in  $\mathcal{B'}_{i,v}$ ($w'$)   &  $[0,\,6]$  & 3 \\
 \rowcolor{nicegreen!20!white} &  Decoding ($K'$) &  $[25,\,500]$   & off  \\ 
\bottomrule
\end{tabular}
\caption{\textbf{Summary of our hyperparameter search for \ours and baselines}. `(S1)' and `(S2)' stand for `Step 1' and `Step 2' as DAB is a 2-step algorithm. $\dagger$ stands for task-adapted algorithms (versus off-the-shelf ones). For adapted versions, we only indicate hyperparameter values that differ from off-the-shelf ones. Evaluations are done for the same number of iterations as in  \autoref{tab:algo_comparison}: N=15 for DAB/DRIV-EX, N=150 for PEZ, N=6 for GCG.  }
\label{tab:hyperparam}
\end{table*}

\section{Appendix: License and terms of use}

The highD dataset \citep{highd} is a public dataset that is free for non-commercial research. We applied for `Access for Non-Commercial Use' via levelxdata's website form\footnote{\url{https://levelxdata.com/highd-dataset/}} and agreed to its terms and conditions, including: using the data for research only, referencing the highD dataset in our paper submission, and only distributing derivative works in as far as they are abstract representations of the dataset. We were granted said access.  

We downloaded the textual transcription of the highD dataset, described in \citet{lc_llm}, that is freely available on the authors' repository\footnote{\url{https://github.com/Pemixing/LCLLM/blob/main/llm_data}} without license specification. 

The data that we used to finetune and evaluate the LLMs contains no personal information as all driving scenarios are anonymized; they only mention vehicle type (`car' or `truck'), speeds, accelerations and positions. Our verification process included parsing the entire train and val sets of the textual highD dataset and verifying that 100\% of the data fell in the exact template shown in \autoref{fig:unbiased_sample}.

Our use of language models (Llama3, Mistral, Qwen2.5, deberta-xlarge) and LoRA finetuning is based on HuggingFace public Hub and libraries. Terms of Service are specified on their website \footnote{\url{https://huggingface.co/terms-of-service}}. 

Our use of the highD dataset and HuggingFace tools is consistent with their intended use, as our sole endeavor was to produce open-source science.

\end{document}